%% file: main.tex
\pdfoutput=1

\documentclass[]{Arxiv/assets/matterlab}

\usepackage{latex/shortcuts}
\usepackage{adjustbox}
\usepackage{ifthen}
\usepackage{tabularx}
\usepackage{algorithm}
\usepackage{algpseudocode}
\usepackage{listings}
\usepackage{xcolor}
\usepackage[inline]{enumitem}

\newcommand{\papertype}{preprint}

\input{Arxiv/assets/shared-preamble.tex}

\title{Schema for In-Context Learning}

\author[1,8]{Pan Chen}
\author[1,*]{Shaohong Chen}
\author[1,*]{Mark Wang}
\author[2,6]{Shi Xuan Leong}
\author[5]{Priscilla Fung}
\author[2,7]{Varinia Bernales}
\author[1,2,3,4,7,8,9,10]{Al\'an Aspuru-Guzik}

\affiliation[1]{\addressCS}
\affiliation[2]{\addressCHEM}
\affiliation[3]{\addressMSE}
\affiliation[4]{\addressCHEMENG}
\affiliation[5]{\addressPSY}
\affiliation[6]{\addressNTU}
\affiliation[7]{\addressAC}
\affiliation[8]{\addressVECTOR}
\affiliation[9]{\addressCIFAR}
\affiliation[10]{\addressNVIDIA}

\abstract{
\input{latex/sections/abstract/abstract}
}

\date{\today}
\correspondence{Al\'an Aspuru-Guzik at \email{alan@aspuru.com}}

\begin{document}
\maketitle

\input{latex/sections/intro/version2}
\input{latex/sections/Related_Work/LiteratureReview}
\input{latex/sections/Methodology/SAICL}
\input{latex/sections/Main_Experiments/exp}
\input{latex/sections/Results/result}
\input{latex/sections/Discussion_and_Future_work/discussion}

\section*{Acknowledgments}
\input{latex/sections/acknowledgement}

{
\small
\bibliography{latex/references}
\bibliographystyle{Arxiv/assets/plainnat}
}

\clearpage
\appendix
\input{latex/sections/appendix/appendix}

\end{document}

%% file: Arxiv/assets/shared-preamble.tex
\usepackage{microtype}
\usepackage{graphicx}
\usepackage{booktabs}
\usepackage{float}
\usepackage{xurl}
\usepackage{hyperref}
\usepackage{titlesec}

\usepackage{amsmath}
\usepackage{amssymb}
\usepackage{mathtools}
\usepackage{amsthm}
\usepackage{bm}

\usepackage[noabbrev,nameinlink]{cleveref}

\input{Arxiv/assets/math_commands}

\usepackage{comment}
\usepackage{longtable}
\usepackage{array}
\renewcommand{\arraystretch}{1.7}
\usepackage{multirow}           
\usepackage{amsfonts}           
\usepackage{nicefrac}
\usepackage{thmtools,thm-restate}
\usepackage{enumitem}
\usepackage{xcolor,colortbl}
\usepackage{caption}
\usepackage{subcaption}
\usepackage{listings}

\newtheorem{theorem}{Theorem}

\newtheorem{definition}[theorem]{Definition}

\newcommand{\addressCHEM}{Department of Chemistry, University of Toronto,  80 St. George St., Toronto, ON M5S 3H6, Canada}
\newcommand{\addressAC}{Acceleration Consortium, 700 University Ave., Toronto, ON M7A 2S4, Canada}
\newcommand{\addressCS}{Department of Computer Science, University of Toronto, 40 St George St., Toronto, ON M5S 2E4, Canada}
\newcommand{\addressVECTOR}{Vector Institute for Artificial Intelligence, W1140-108 College St., Schwartz Reisman Innovation Campus, Toronto, ON M5G 0C6, Canada}
\newcommand{\addressPSY}{Department of Pyschology, University of Toronto, Sidney Smith Hall, 100 St. George Street, ON M5S 3G3, Toronto, Canada}
\newcommand{\addressNTU}{School of Chemistry, Chemical Engineering and Biotechnology, Nanyang Technological University, Singapore, Singapore}
\newcommand{\addressMSE}{Department of Materials Science \& Engineering, University of Toronto, 184 College St., Toronto, ON M5S 3E4, Canada}
\newcommand{\addressCHEMENG}{Department of Chemical Engineering \& Applied Chemistry, University of Toronto, 200 College St., Toronto, ON M5S 3E5, Canada}

\newcommand{\addressCIFAR}{Canadian Institute for Advanced Research (CIFAR), 661 University Ave., Toronto,
ON M5G 1M1, Canada}
\newcommand{\addressNVIDIA}{NVIDIA, 431 King St W \#6th, Toronto, ON M5V 1K4, Canada}

\providecommand{\doi}[1]{\href{https://doi.org/#1}{doi:#1}}


%% file: Arxiv/assets/math_commands.tex
\usepackage{amsmath,amsfonts,bm}

\def\eqref#1{equation~\ref{#1}}

\def\1{\bm{1}}

\DeclareMathAlphabet{\mathsfit}{\encodingdefault}{\sfdefault}{m}{sl}
\SetMathAlphabet{\mathsfit}{bold}{\encodingdefault}{\sfdefault}{bx}{n}



%% file: latex/sections/abstract/abstract.tex
While {\ICLFull} ({\ICL}) enables {\LLMsFull} ({\LLMs}) to perform tasks without parameter updates, models often rely on surface-level shortcuts rather than abstract reasoning. We attribute this limitation to {\SchemaDormancy} — the failure to spontaneously activate latent reasoning frameworks. 
To address this, we introduce Schema-Activated {\ICL} ({\sysname}). Motivated by {\SchemaTheory}, {\sysname} explicitly activates an abstracted schema before the query, decoupling reasoning structure from task content to enforce structural mapping.
We evaluate {\sysname} across the domains of {\science}, {\stat}, and {\finance} using seven diverse models, including GPT-4o and {\deepseek}. Experiments stratified by knowledge similarity reveal that {\sysname} consistently outperforms {\OneShotCoT} baselines, particularly in high-knowledge-density scenarios. 
Furthermore, {\sysname} demonstrates superior robustness against retrieval noise and domain-incongruent examples. 
Finally, we quantify the semantic drift between latent and activated schemas, providing insights into the correlation between "Schema Shift" and downstream performances.

%% file: latex/sections/intro/version2.tex
\section{Introduction}
\subsection{Learning in humans and LLMs shares similar underlying patterns}
{\ICLFull} ({\ICL}) has emerged as a dominant approach for adapting {\LLMsFull} ({\LLMs}) to new tasks without requiring fine-tuning or additional parameter updates. 
By conditioning on demonstrations, {\ICL} enables LLMs to leverage prior knowledge and generalize to unseen examples. 
Despite its effectiveness, cognitively, traditional {\ICL} is not fully aligned with how humans acquire and apply knowledge in real-world learning scenarios, as it lacks mechanisms for episodic memory and context-rich encoding \citep{li2024linking}.

\textbf{Learning in humans} is inherently structured and involves knowledge abstraction, retrieval, and adaptive reasoning. Research in cognitive science suggests that humans develop mental frameworks, called \textit{schemas}, that organize prior knowledge and facilitate problem solving in new contexts \citep{rumelhart1977schemata}. 
These schemas enable efficient retrieval of relevant information and guide interpretation and action, reducing reliance on explicit demonstrations \citep{rumelhart1977schemata}. 
Critically, \textbf{Schema Activation}, which consists of bringing the proper schema into working memory, is essential for effective comprehension and analogical transfer; retrieval alone may not suffice \cite{gick1983schema,gentner1983structure}. Failure to possess or activate a schema might result in limited comprehension of the new information \citep{bransford1972contextual}.

\textbf{Learning in LLMs} mirrors this limitation. For example, recent models such as GPT-4 retrieve numerous plausible analogs with high recall but often select incorrect ones because they rely on surface-level similarity rather than structural alignment \citep{puranam2025analogical}. This setback calls for mechanisms that go beyond retrieval, mobilizing schema-like abstractions to guide reasoning. We consider a failure or absence of schema activation a phenomenon of \textbf{{\SchemaDormancy}} (\cref{def:schema_dormancy}).

\begin{tcolorbox}[
    colback=white, 
    colframe=black, 
    sharp corners, 
    boxrule=0.5pt, 
    fonttitle=\bfseries
]
\begin{definition}[{\SchemaDormancy}]
\label{def:schema_dormancy}
\textit{\textbf{{\SchemaDormancy}} is the phenomenon where a large language model fails to spontaneously activate a relevant abstract reasoning template (schema) during inference, despite demonstrating the capacity to utilize that schema when explicitly prompted.}
\end{definition}
\end{tcolorbox}

In a cognitive science experiment in 1972 \citep{bransford1972contextual}, humans' ability to memorize a messy paragraph was enhanced when a schema was activated. 
With a large language model, we repeated\footnote{The original study provided a seven-word context ('this paragraph is all about washing clothes'), whereas our study used the condensed prompt: 'Washing Machine'.} this famous experiment and found that, as reported in \cref{sec:washing_machine}, with the exact two words \textit{{\WashingMachine}}, the attention layers changed significantly. {\SchemaTheory} contributes to comprehension in psycholinguistics and provides insights into how LLMs interpret information computationally.

\subsection{{\SchemaTheory} can be conceptualized and implemented with LLMs}
Although the {\WashingMachine} experiment demonstrates that a concise cue can induce measurable changes in LLMs' attention dynamics, in many real-world scenarios, such as scientific, mathematical, and financial contexts, it may not always be feasible to provide a shorter, more precise hint to activate the schema. Therefore, we conceptualize schema-based thinking as a set of explicit steps, namely \textbf{REAP}: Problem Representation (R), Prior Schema Retrieval (R), Episodic Example Retrieval (E), and Schema Activation (A). This retrieval and activation process enables LLMs to solve problems more efficiently by integrating structured prior knowledge into their current reasoning, addressing the structural mapping gap observed in LLM analogical reasoning \citep{puranam2025analogical}. Echoing approaches using latent graph schemas for fast transfer learning \citep{guntupalli2023graph}, our method enforces structured abstraction as the medium for retrieval, reasoning, and inference.

We evaluated {\sysname} on three main benchmarks: GPQA-Diamond ({\science}), MMLUProStat ({\stat}), and MMLUProFinance ({\finance}) \citep{rein2024gpqa, wang2024mmlu}, and one for multi-modal: LabSafety \citep{zhou2024labsafety}. The results show that the use of {\sysname} improves accuracy over standard {\ICL} (\oneshot) in most scenarios, especially when the knowledge similarity is high. 
Our results show that traditional one-shot prompting, whether or not a reasoning path is provided, may fail tasks due to {\SchemaDormancy}. 

%% file: latex/sections/Related_Work/LiteratureReview.tex
\section{Related Work}
\subsection{Schema From the lens of Cognitive Psychology}
{\SchemaTheory} is one of the fundamental theories of cognitive psychology. This theory conceptualizes how humans organize and structure knowledge into coherent mental frameworks, or schemas, constructed from prior experiences and serving as interpretive structures for understanding new information \cite{rumelhart1977schemata}. These abstract structures are dynamic; they actively guide how prior knowledge is encoded and retrieved and how the new information is perceived \citep{brewer1981role}. Foundational research in psychology from decades ago, such as \cite{bartlett1932remembering} and \cite{piaget1952origins}, established that human learning involves either interpreting new information within existing schemas (\textit{assimilation}) or modifying these schemas to incorporate novel knowledge (\textit{accommodation}). Activated schemas from the lens of prior knowledge provide a cognitive mechanism that enables efficient problem-solving and reasoning by guiding the retrieval process and allowing individuals to make inferences and fill in missing details \citep{anderson1978recall, piaget1952origins}. This model of human cognition, in which activating the correct abstract structure is key to interpreting a new problem, provides direct theoretical motivation for the {\sysname} framework.

\subsection{{\ICLFull} Methodologies}
\subsubsection{Conventional ICL Methods}
{\oneshot} and few-shot learning paradigms \citep{NEURIPS2020_1457c0d6} have been adopted as computationally efficient methodologies \citep{Parnami2022LearningFF}. Example-driven ICL utilizes predefined question-answer pairs to change the probability distribution of output tokens conditioned on user queries and prior knowledge \citep{10.1145/3386252, Min2022RethinkingTR}. Despite state-of-the-art (SOTA) LLMs achieving substantially longer context windows, few-shot learning continues to demand substantial computational resources as the number of demonstration samples and their associated token counts increase, resulting in inevitable computational cost inflation \citep{keles2022}. Furthermore, LLMs demonstrate sensitivity to performance worsening when processing long in-context demonstrations for complex reasoning tasks, making example-driven ICL a compromise for tasks characterized by complex reasoning \citep{li2024long}.

Chain-of-Thought (CoT) reasoning has arisen as a critical strategy within ICL by asking LLMs to explicitly outline intermediate steps before arriving at a final answer, to enhance the reasoning capabilities \citep{wei2022chain, kojima2022large}, thus significantly improving LLM performance on tasks that require multi-step inference \citep{nye2021show, wang2022self}. However, CoT reasoning usually operates within example-driven reasoning frameworks, where explicit reasoning details are provided through a few demonstrations to guide model outputs \citep{zhang2022automatic}. Recent studies on CoT prompting include generating multiple thought paths for reasoning before arriving at a final answer \citep{wang2022self}, and progressively refining reasoning paths by starting with simpler sub-questions \citep{zhou2022least}.

Existing ICL research conceptualizes LLMs primarily as pattern-matching systems operating over prompt examples without comprehensive analysis of their internal abstraction mechanisms \citep{min2021metaicl, levine2022parallel}, or as instance-specific systems wherein the model constructs novel linear reasoning sequences tailored to each input query, consequently necessitating substantial human engineering of input queries to achieve satisfactory performance on specific tasks \citep{stechly2024cot}. {\sysname} advances the exploration of how language models can generate reasoning processes through their internal knowledge representations in a human-interpretable manner, simultaneously optimizing both performance and interpretability.

\subsubsection{Abstraction-Driven Approaches}

While most existing {\ICL} methods were heavily example-driven, prior work has raised key issues \citep{saglam2025learning, lampinen2024broader, dong2022survey}. Recently, the machine learning (ML) community has witnessed growing development of {\ICL} approaches in a broader sense \citep{lampinen2024broader}, including abstraction-driven {\ICL} (A-ICL) \citep{swaminathan2023schema}, which help understand how models understand and utilize context. Although previous works have linked the mechanism of induction heads in LLMs to the contextual maintenance and retrieval (CMR) model in human episodic memory \citep{olsson2022inductionheads, polyn2009cmr}, direct evidence for high-level schema induction in language models remains limited.

Limitations and concerns have been raised regarding the similarities and differences between LLMs and humans in terms of internal knowledge understandings in the scope of abstract driven ICL, especially regarding conflicts between internal knowledge and context \cite{obiso-etal-2025-dynamic}, whether LLMs could reason in an abstract level \cite{yun2025abstract}, and whether LLMs understand individual lexicons in the same way as humans (Human-Likeliness).

In contrast, our work adopts another A-ICL approach that explicitly extracts general reasoning steps and conducts experiments at a higher level than task-specific knowledge. Prior studies have shown that fixed, structured generation — enabled by carefully designed prompting mechanisms or properly constrained decoding strategies — can improve LLMs' performance on reasoning tasks (e.g., ReAct, Program-of-Thoughts) \citep{yao2023react, chen2023pot}. 
Furthermore, retrieval of previous LLM-generated schemas or demonstrations has been shown to be autoregressively beneficial for ICL, as in prompt-retrieval and retrieval-augmented methods \citep{rubin2022retrieveprompts, shi2024replug}.

\subsection{Other Human-Inspired Prompting Methods}
Existing human-inspired {\ICL} approaches include Metacognitive Prompting \cite{metacognitive}, Thread of Thought \cite{thread_of_thought}, and Least-to-Most (LtM) \cite{zhou2022least}. Yet, there is no unifying framework grounded in cognition to explain and facilitate these strategies. 
Retrieval-Augmented Generation (RAG) provides LLMs with access to prior knowledge within a given knowledge base for future queries on similar tasks, which can be considered long-term memories for LLMs. However, traditional RAG limits the quality of the retrieval strategy and the corresponding reasoning logic learned from prior-knowledge examples to the quality of the existing knowledge base examples. It remains a challenge for existing RAG techniques to adapt to dynamic and interconnected knowledge bases \cite{gutierrez2025rag}.
\textbf{{\sysname}} is built on the RAG paradigm and leverages {\SchemaTheory} to unify prompting methodologies through a cognitive science lens.

%% file: latex/sections/Methodology/SAICL.tex
\section{Motivating and Preliminary Studies}
\textbf{In our motivating study}, we fed the ambiguous passage (\Cref{app:washingmachine}) describing laundry workflow (the \textit{{\WashingMachine}} passage) to Llama-3.1-8B in BF16 precision under two conditions: \textit{Dormant State} (no context) and \textit{Activated State} (context provided).

Attention Layer Analysis: We reported the self-attention weights from the attention layers. In summary, we observed:
\begin{itemize}
    \item \textit{Dormant State (Diffuse Attention)}: Attention heads exhibited high entropy. Tokens like "groups" and "materials" attended primarily to immediate syntactic neighbours. The model failed to resolve co-references (e.g., linking "procedure" to "arrange").   
    \item \textit{Activated State (Focused Structural Attention)}: Upon injecting the \textit{{\WashingMachine}} keyword, attention entropy collapsed. Specific induction heads routed strong weights from the schema keyword to ambiguous terms, effectively grounding abstract concepts.
\end{itemize}
\input{latex/figure_tables/attention_entropy}
This experiment confirms that schema activation is a tangible mechanical phenomenon: a keyword can restructure the flow of information from surface-level association to structural coherence.

\label{sec:washing_machine}
\textbf{In our preliminary study}, we observed that, compared with traditional {\OneShot}, the models would perform better with improvements up to 36\% in certain scenarios (\Cref{app:preliminary}) if explicitly asked to go through the \textbf{REAP} process first. Moreover, we observed that this process led to fewer tokens in the last layer of the workflow (\Cref{app:interpretability_exps}).

These motivating and preliminary studies strongly encourage investigations into how to tackle {\SchemaDormancy} in more complex scenarios and into building a workflow that faithfully mimics humans' cognitive processes when needed.

\section{{\sysnamefull}}

We propose this innovative ICL framework, {\sysnamefull} (\sysname), which mimics how humans use prior examples to activate a schema, enabling better understanding and solutions to new problems. 
This framework is simple, yet flexible and unified with other {\ICL} strategies, providing a reliable and transparent explanation of how a language model learns from previous examples and turns these abstractions into a more powerful schema that guides it in solving novel problems.

\subsection{Overall Workflow}

\begin{figure*}[h]
  \centering
  \includegraphics[width=\linewidth]{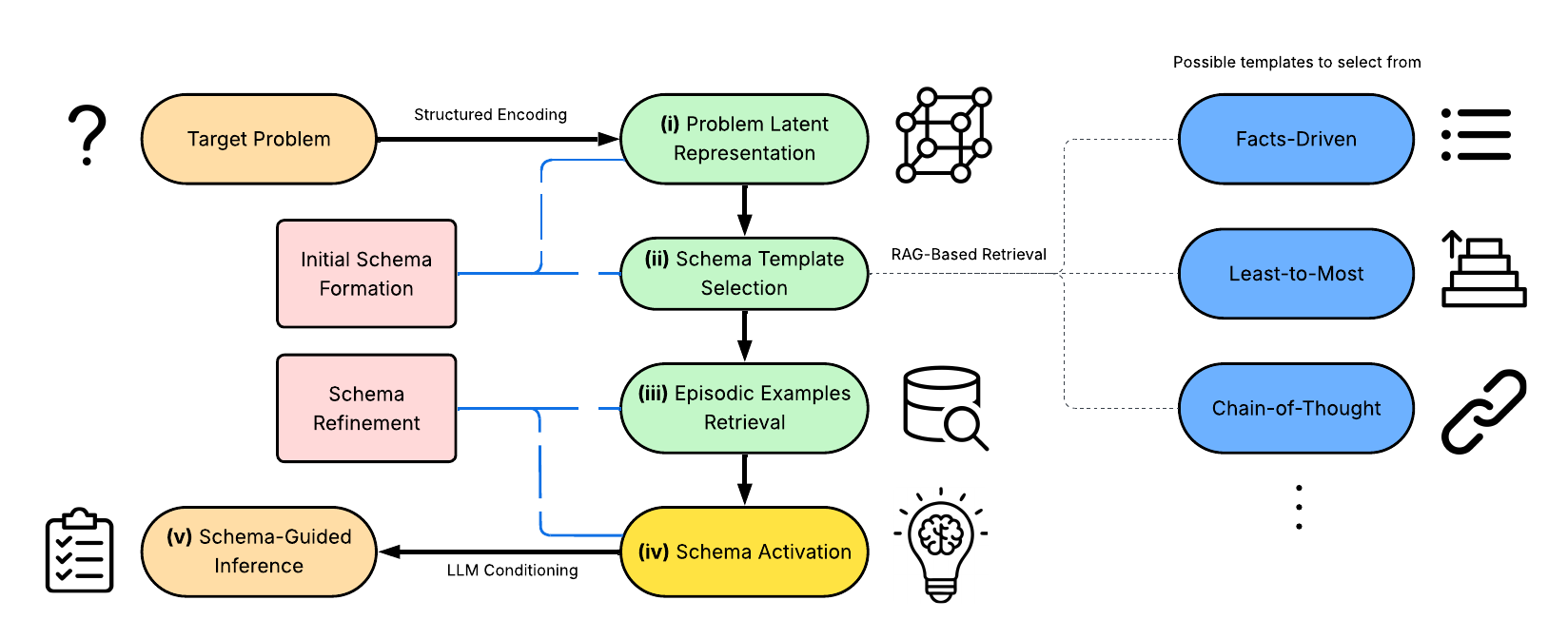}
  \caption{{\sysname}
  \label{fig:idea}general workflow. The framework demonstrates five sequential stages: \textbf{(i)} takes a target problem as input and generates its corresponding latent representation, \textbf{(ii)} utilizes a flexible schema template (e.g., Least-to-Most, Chain-of-Thought, etc.) to form an initial schema and identify relevant properties of the question, \textbf{(iii)} gathers a subset of past influential episodic examples that are relevant to solve the target problem, \textbf{(iv)} performs schema activation and integrates retrieved knowledge to develop a refined schema to solve the target problem, and \textbf{(v)} conditions the target problem to the adapted schema and utilizes schema-guided inference to reach a solution.}
\end{figure*}

\input{latex/sections/Methodology/algorithm}

{\sysname} operationalizes {\SchemaTheory} from cognitive science in five steps, aligning abstract schema formation with language model retrieval and reasoning.

\ifthenelse{\equal{\papertype}{preprint}}{
    \begin{enumerate}[label=(\roman*)]
        \item \textbf{Problem Latent Representation}:  
        Given an input problem $x$, the LLM constructs a representation $\mathcal{S}_x = \mathcal{R}(x)$. This representation functions as the initial schema for the new problem.
        
        \item \textbf{Prior Schema Retrieval and Selection}:  
        {\sysname} retrieves the most relevant schema $\hat{\mathcal{S}} \in \mathcal{S}$ that maximizes similarity with $\mathcal{S}_x$. {\sysname} is compatible with various schema templates (e.g., Least-to-Most, Chain-of-Thought, etc.).
        
        \item \textbf{Episodic Examples Retrieval}:
        Conditioned on the retrieved schema $\hat{\mathcal{S}}$, {\sysname} collects a subset of episodic examples whose decayed association weights $w_{\hat{\imath} j}(t)$ exceed a threshold $\tau$. This yields a set $\hat{\mathcal{E}}_\tau$ of examples that remain influential for the current reasoning.
        
        \item \textbf{Schema Activation (Assimilation / Accommodation)}:
        The retrieved schema $\hat{\mathcal{S}}$ and episodic set $\hat{\mathcal{E}}_\tau$ are integrated with the current problem representation, producing a new activated schema: $\mathcal{S}_{\text{new}} = f(\mathcal{S}_x, \hat{\mathcal{S}}, \hat{\mathcal{E}}_\tau)$.
        
        This integration may proceed through \textit{assimilation} when the prior schema fits well, or \textit{accommodation} when internal restructuring is required.
        
        \item \textbf{Schema-Guided Inference}: 
        Finally, the LLM solves the task by conditioning on the input $x$ and the adapted schema $\mathcal{S}_{\text{new}}$: $y = \text{LLM}(x, \mathcal{S}_{\text{new}})$.
    \end{enumerate}
}{
    \begin{enumerate*}[label=(\roman*)]
        \item \textbf{Problem Latent Representation}:  
        Given an input problem $x$, the LLM constructs a representation $\mathcal{S}_x = \mathcal{R}(x)$. This representation functions as the initial schema for the new problem.
        
        \item \textbf{Prior Schema Retrieval and Selection}:  
        {\sysname} retrieves the most relevant schema $\hat{\mathcal{S}} \in \mathcal{S}$ that maximizes similarity with $\mathcal{S}_x$. {\sysname} is compatible with various schema templates (e.g., Least-to-Most, Chain-of-Thought, etc.).
        
        \item \textbf{Episodic Examples Retrieval}:
        Conditioned on the retrieved schema $\hat{\mathcal{S}}$, {\sysname} collects a subset of episodic examples whose decayed association weights $w_{\hat{\imath} j}(t)$ exceed a threshold $\tau$. This yields a set $\hat{\mathcal{E}}_\tau$ of examples that remain influential for the current reasoning.
        
        \item \textbf{Schema Activation (Assimilation / Accommodation)}:
        The retrieved schema $\hat{\mathcal{S}}$ and episodic set $\hat{\mathcal{E}}_\tau$ are integrated with the current problem representation, producing a new activated schema: $\mathcal{S}_{\text{new}} = f(\mathcal{S}_x, \hat{\mathcal{S}}, \hat{\mathcal{E}}_\tau)$.
        
        This integration may proceed through \textit{assimilation} when the prior schema fits well, or \textit{accommodation} when internal restructuring is required.
        
        \item \textbf{Schema-Guided Inference}: 
        Finally, the LLM solves the task by conditioning on the input $x$ and the adapted schema $\mathcal{S}_{\text{new}}$: $y = \text{LLM}(x, \mathcal{S}_{\text{new}})$.
    \end{enumerate*}
}

\Cref{fig:idea} describes the conceptual pipeline of {\sysname}. \Cref{app:algorithm_walkthrough} details the complete mathematical formalization of each step in the framework. \Cref{alg:sa-icl} summarizes {\sysname} framework. \Cref{app:full_conversation_solver} shows the example of using the activated schema at inference.

%% file: latex/figure_tables/attention_entropy.tex

\ifthenelse{\equal{\papertype}{preprint}}{
    \begin{figure}[htpb]
      \centering
      \includegraphics[width=0.49\linewidth]{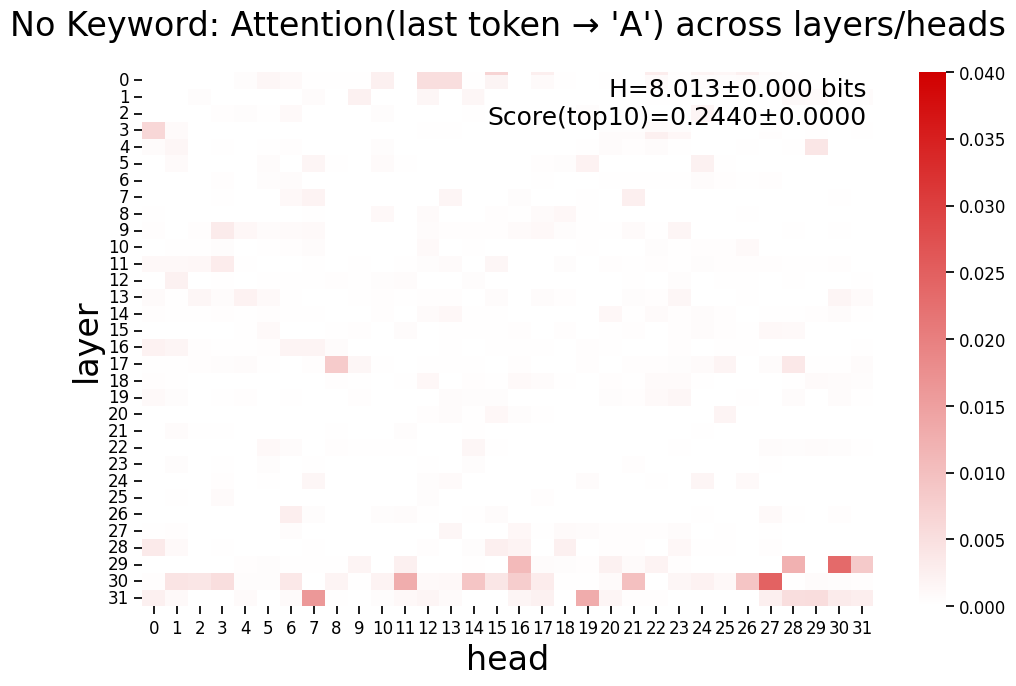}
      \hfill
      \includegraphics[width=0.49\linewidth]{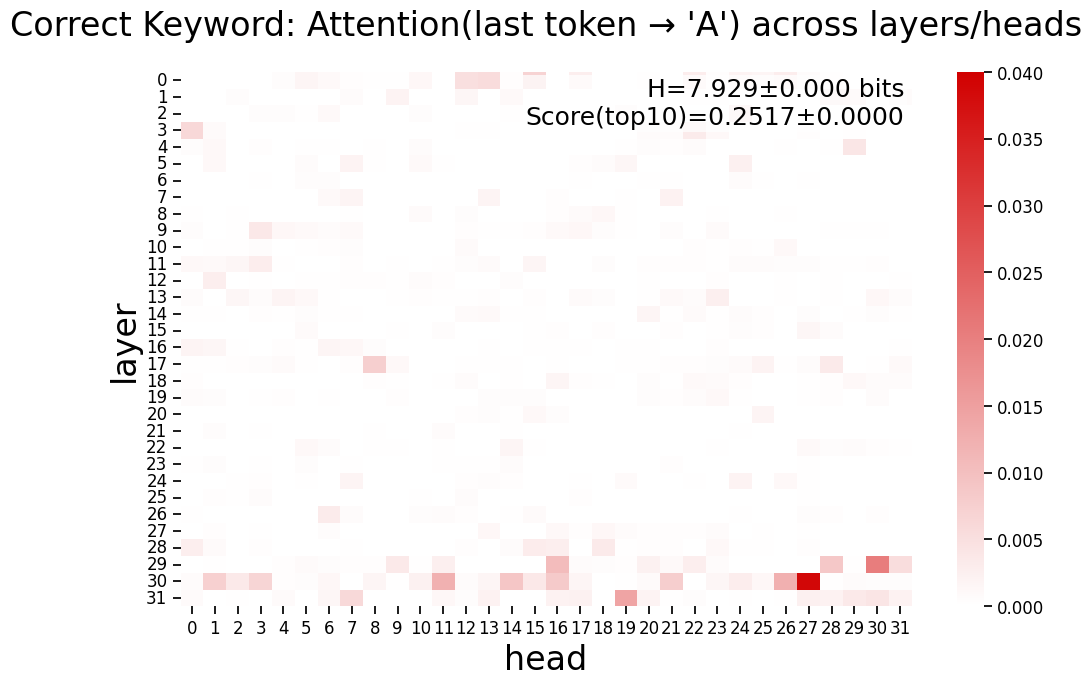}
      \caption{By activating a schema, attention entropy is reduced. In particular, induction heads were more diverse without a correct activation keyword, indicating a stronger awareness of the details under the grounding abstract concepts.}
    \end{figure}
}{
    \begin{figure}[htpb]
      \centering
      \includegraphics[width=\linewidth]{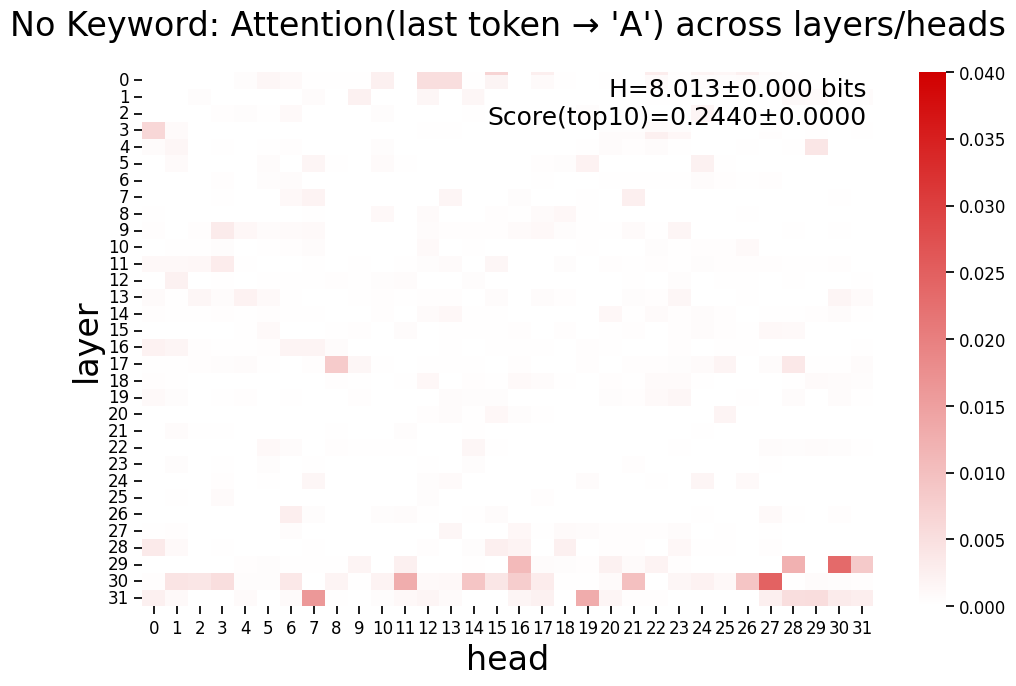}
      \includegraphics[width=\linewidth]{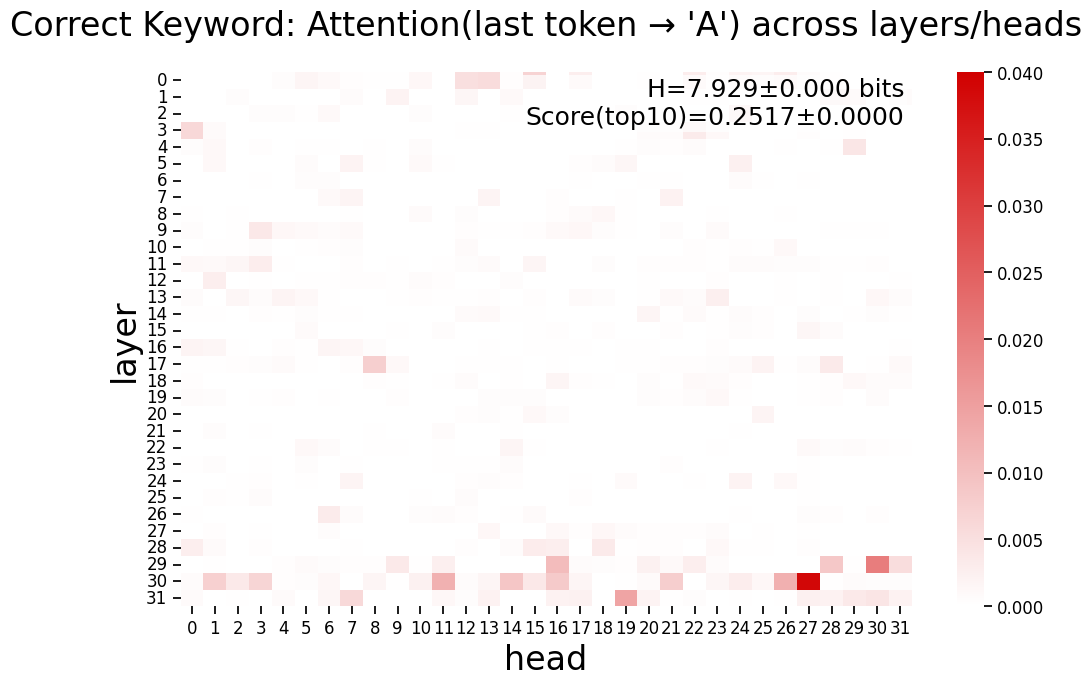}
      \caption{By activating a schema, attention entropy is reduced. In particular, induction heads were more diverse without a correct activation keyword, indicating a stronger awareness of the details under the grounding abstract concepts.}
    \end{figure}
}

%% file: latex/sections/Methodology/algorithm.tex
\begin{algorithm}[h]
\caption{\sysnamefull}
\label{alg:sa-icl}
\small
\begin{algorithmic}[1]
\Require Problem $x$; Schemas $\mathcal{S}=\{\mathcal{S}_1,\dots,\mathcal{S}_N\}$; Episodic set $\mathcal{E}=\{e_1,\dots,e_M\}$; Memory $\mathcal{M}=\{(\mathcal{S}_i,e_j,w_{ij}(t))\}$; Representation $\mathcal{R}$; Similarity $\mathrm{sim}$; Activation $f$; $\mathrm{LLM}$; Threshold $\tau \in [0,1]$
\Ensure $y$, $\mathcal{S}_{\mathrm{new}}$

\State $\mathcal{S}_x \gets \mathcal{R}(x)$
\State $\hat{\imath} \gets \arg\max_{i \in \{1,\dots,N\}} \; \mathrm{sim}(\mathcal{S}_x, \mathcal{S}_i)$
\State $\hat{\mathcal{S}} \gets \mathcal{S}_{\hat{\imath}}$
\State $\hat{\mathcal{E}}_{\tau} \gets \emptyset$
\For{$j \in \{1,\dots,M\}$}
    \If{$w_{\hat{\imath} j}(t) \ge \tau$}
        \State $\hat{\mathcal{E}}_{\tau} \gets \hat{\mathcal{E}}_{\tau} \cup \{e_j\}$
    \EndIf
\EndFor
\State $\mathcal{S}_{\mathrm{new}} \gets f(\mathcal{S}_x, \hat{\mathcal{S}}, \hat{\mathcal{E}}_{\tau})$
\State $y \gets \mathrm{LLM}(x, \mathcal{S}_{\mathrm{new}})$
\State \Return $(y, \mathcal{S}_{\mathrm{new}})$

\end{algorithmic}
\end{algorithm}

%% file: latex/sections/Main_Experiments/exp.tex
\section{Main Experiments}
\subsection{Technical Setup}
All experiments used standardized OpenAI-style API endpoints. For open-source models, we used an NVIDIA A40 GPU with 24GB VRAM. \Cref{app:experiment_setup} summarizes the model families we used and the corresponding execution environments.

\subsection{Tasks and Metrics}

\textbf{Tasks.} We designed closed-ended multiple-choice question-answering tasks. The experimental design incorporated multiple knowledge density levels to evaluate model performance across varying degrees of prior knowledge acquisition.

\textbf{Datasets.} The experiments primarily utilized GPQA-Diamond ({\science}), MMLUProStat ({\stat}), and MMLUProFinance ({\finance}), three rigorously annotated benchmarks containing questions across five domains. These datasets serve as the questions and knowledge base for both schema formation and example retrieval. To have a larger knowledge base to simulate different levels of real-world knowledge density, GPT-4o was employed to generate three synthetic variants for each problem. The criteria defined three distinct similarity levels: {\same}, {\similar}, and {\different}. The resulting datasets in the synthetic retrieval pool are labelled \textit{synthetic}. Detailed prompt templates we used to generate the synthetic retrieval pool are provided in \Cref{app:synthetic_data_generation}. An analysis for knowledge density of the synthetic examples is reported in \Cref{app:knowledge_density}.

\textbf{Baselines.} We compared {\sysname} with {\OneShot} and {\OneShotCoT}. For fairness of comparisons, we ensured that all baselines received the same example as {\sysname}. 

\textbf{Schema Template.} To ensure fair comparisons, we implemented a template of CoT in our main experiments. This also serves as an illustrative example to demonstrate how {\sysname} can naturally integrate with existing prompting methods within a unified framework. In our ablation studies, we integrated {\sysname} with four other schema templates.

\subsection{Different Experiments}
\textbf{Main Experiments}: In our main experiments, we tested all seven LLMs across the three datasets at three knowledge level densities (\textit{Essentially Same}, \textit{Similar}, \textit{Different}). To reduce noise and ensure fair comparison, we used synthetic data as the retrieved knowledge, yielding 63 distinct settings.

\textbf{Effectiveness of activations}: We tested the effectiveness of schema activation with four other schema templates, including Least-to-Most \citep{zhou2022least}, Step-Back \citep{Zhengetal2023}, Deductive-driven, and Facts-driven, with {\deepseek} and Gemini 2.0-Flash. We wanted to see whether the effectiveness was limited to CoT prompting. The schema templates were reported in \Cref{app:schema_templates}

\textbf{Loss of Relevant Examples}: In this experiment, we asked all seven models to complete questions from the GPQA-Diamond (Science) dataset, excluding the biology subset, since the biology subset has the fewest questions, and used one of the biology questions as the in-context example. This simulated a scenario in which a mistake was made during the retrieval process.

\textbf{End-to-end studies}: We conducted two end-to-end studies. (1) An example was chosen by a {\cohere} conditioned on the schemas of prior knowledge and the problem itself. We wanted to investigate the feasibility of deploying {\sysname} more closely to the real world. (2) {\deepseek} was asked to self-generate an example, similar to Analogy Prompting \citep{bhavya-etal-2022-analogy}.

\textbf{Multi-modal dataset}: Humans often generate schemas from their visual perception, therefore we tested {\sysname} with Gemini 2.0-Flash on LabSafety. We reported the result in \Cref{app:vision}. 

%% file: latex/sections/Results/result.tex
\section{Main Results}


\subsection{{\SchemaDormancy} is present even in high knowledge density scenarios}
Language models achieved moderate accuracy when provided only with high-quality examples, yet their performance remained suboptimal compared to models that employ explicit schema-based learning techniques (\Cref{fig:knowledge_gap}). This performance gap suggests that relying solely on models' internal representations for high-level abstraction fails to fully exploit the learning potential of quality demonstrations (\Cref{def:schema_dormancy}). Such results mirror the findings from cognitive psychology, where comprehension suffered not from missing information but rather from the absence of an activated schema to effectively organize the context \cite{bransford1972contextual}. When models were conditioned to implement schema-based learning through structured guidelines and templates (\Cref{alg:sa-icl}), we consistently observed significant improvements in downstream task performance, indicating the existence of {\SchemaDormancy} when the LLMs are not explicitly prompted to activate their refined schemas. These performance gains were consistent even when using domain-agnostic schemas.

\begin{figure*}[t]
  \centering
  \includegraphics[width=\linewidth]{figure/main_performance_combined_same_similar.pdf}
  \caption{The number indicates the difference between {\sysname} and the best alternative. 
  Despite being given the same example, {\sysname}, {\OneShotCoT} does not reach the optimal performance, often perform same or even worse than {\oneshot}. On the other hand, regardless of whether the example is \textit{Essentially Same} or \textit{Similar}, {\sysname} (Blue) provides a better reasoning boost compared to the baselines in most cases.}
  \label{fig:knowledge_gap}
\end{figure*}

Nevertheless, {\sysname} is considered remarkably effective. On average, across all knowledge levels, the improvement is +5.1\% (\textbf{$p$-value} < 0.001 ***\footnote{The number of "*"s indicate the decimal precision.}). 
Analysis per model was reported in \Cref{app:extra}, indicating {\sysname} is effective for five out of seven models. 
\Cref{tab:p_value_per_dataset} shows {\sysname} as an effective methodology for all of the three domains we tested. 



\input{latex/sections/Results/pvalue_per_dataset}

    

\subsection{Density of knowledge affects the performance of {\sysname}}
The experimental results presented in \Cref{fig:trend} demonstrate a stratified performance profile for the {\sysname} methodology. Under optimal conditions, when the provided exemplar exhibits essential equivalence to the target question, the {\sysname} approach achieves complete dominance with a perfect win rate against the {\oneshot} baseline. In sub-optimal conditions characterized by varying degrees of quality and similarity, the {\sysname} method maintains consistent performance advantages, demonstrating superior results across the majority of experimental conditions even when exemplar-question alignment is imperfect.

We observe a trend that, as the knowledge level drops, although still outperforming the baseline, the effectiveness of {\sysname} also drops. This behaviour is well aligned with {\SchemaTheory} from a cognitive science perspective.

\input{latex/figure_tables/trend}

These findings indicate that while {\sysname} achieves maximum efficacy when provided with highly relevant exemplars, its performance benefits extend beyond ideal matching conditions. The evidence supports the conclusion that {\sysname} constitutes a fundamentally superior prompting methodology rather than one dependent solely on exceptional example quality, thereby establishing its viability as a generalizable improvement to existing in-context learning frameworks.



\subsection{Ablation Study: Effectiveness of the Schema Activation step}

\ifthenelse{\equal{\papertype}{preprint}}{
    \begin{figure}[t]
      \centering
      \includegraphics[width=0.65\linewidth]{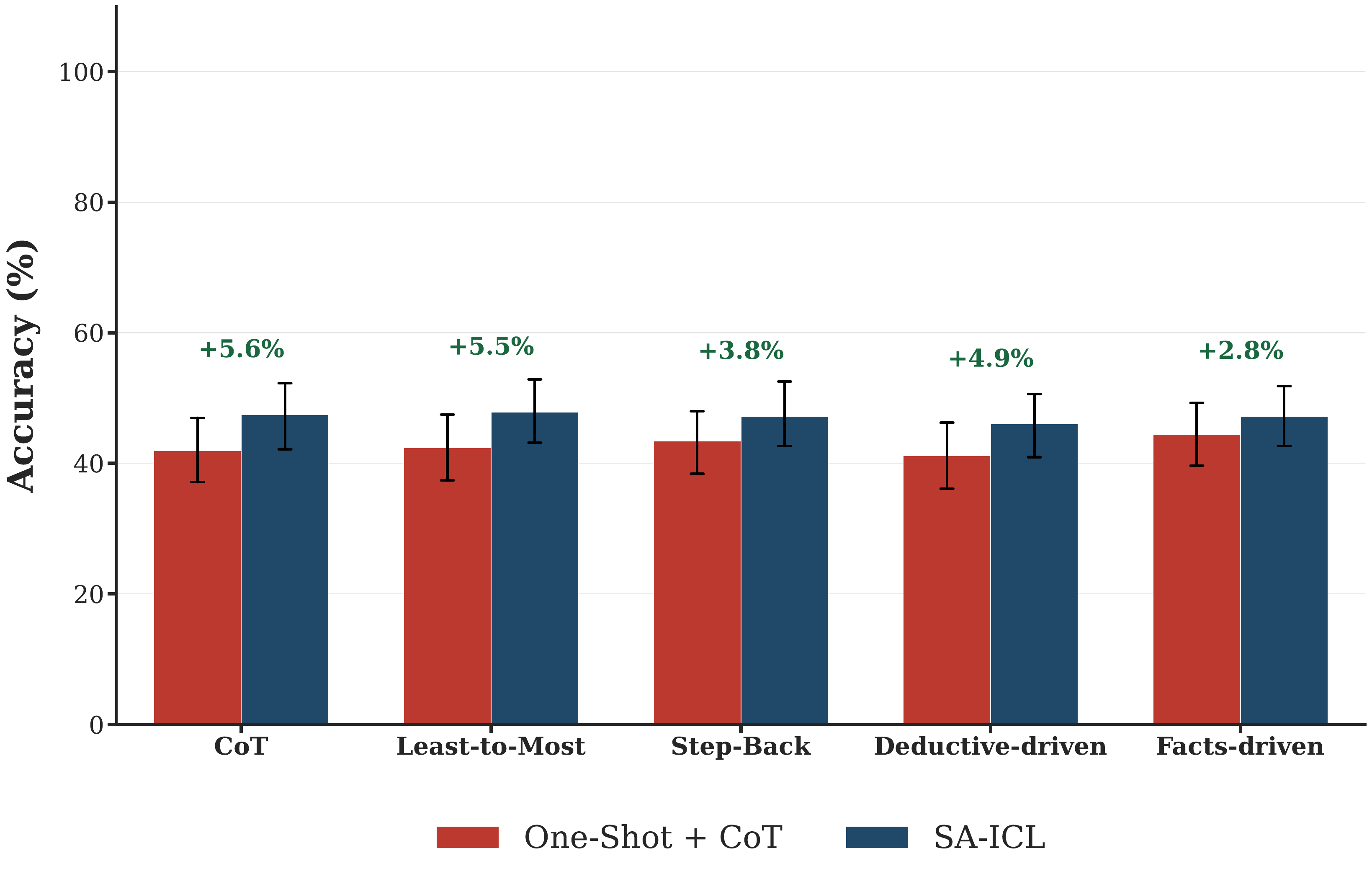}
      \caption{Evaluation of four different interpretable schema templates using {\deepseek} and Gemini on the {\science} questions (with Similar examples). {\sysname} demonstrates a statistically significant improvement in the adoption of all five templates with a global $p$-value = 0.0044 (**)}
      \label{fig:templates}
    \end{figure}
}{
    \begin{figure}[t]
      \centering
      \includegraphics[width=\linewidth]{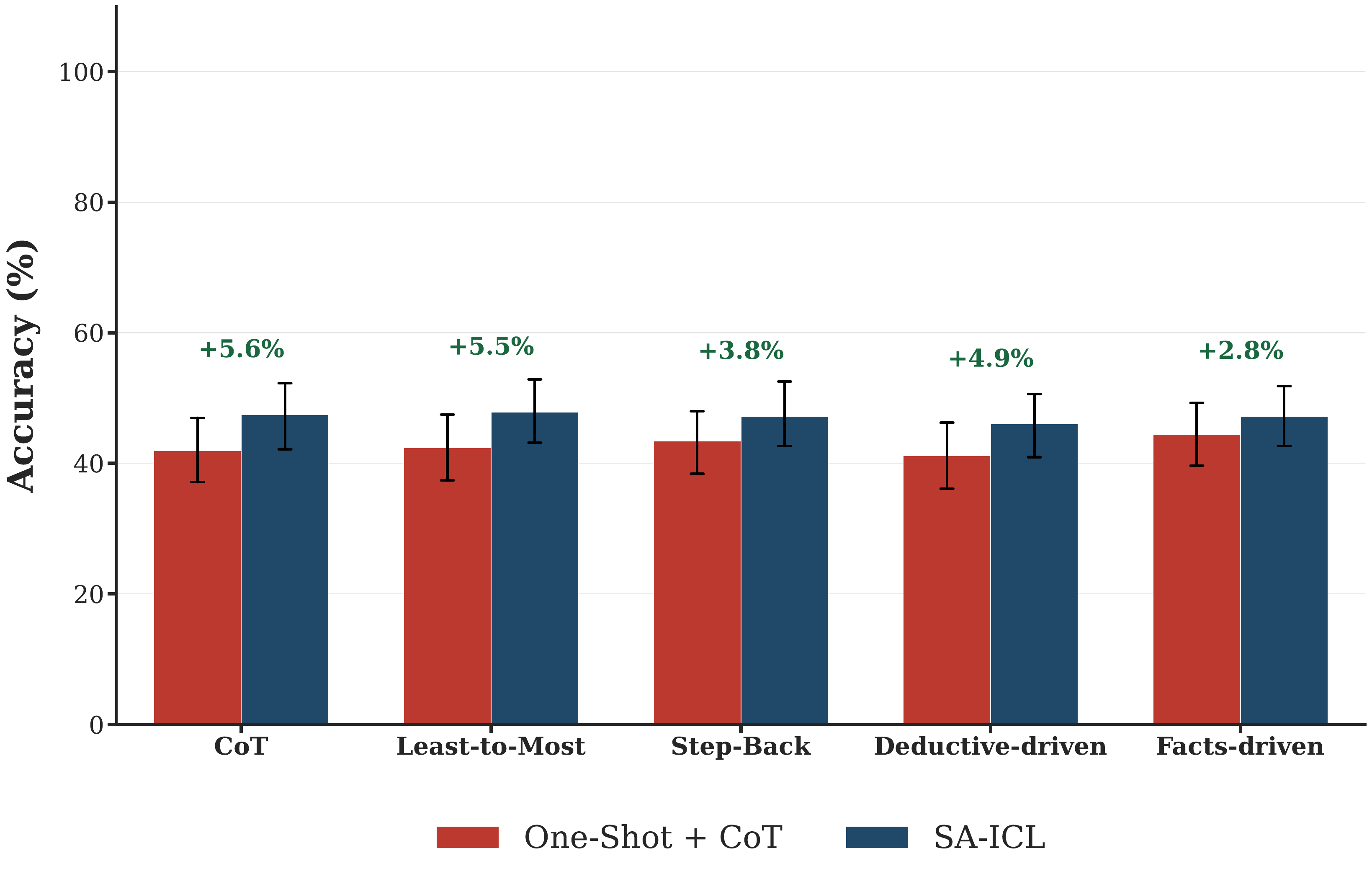}
      \caption{Evaluation of four different interpretable schema templates using {\deepseek} and Gemini on the {\science} questions (with Similar examples). {\sysname} demonstrates a statistically significant improvement in the adoption of all five templates with a global $p$-value = 0.0044 (**)}
      \label{fig:templates}
    \end{figure}
}

Across five schema templates, we observed a consistent improvement in {\sysname} over one-shot + CoT (\Cref{fig:templates}). This is a strong and promising signal that, despite the different nature of schema templates, having them unified under {\sysname} helped the models interpret the context better. This finding bridges the shared underlying thinking patterns between LLMs and humans, highlighting the importance of explicitly activating human-like schemas when solving complex problems.

\subsection{Ablation Study: Loss of relevant knowledge and Schema Disruption}
When giving LLMs a biology example in the context of solving chemistry and physics questions, we observed weaker performance from {\sysname} (\Cref{app:history}), which aligns with the trend we observed in \Cref{fig:trend}. When the knowledge is of \textit{Different} level and thus irrelevant, it is clear that \hyperref[def:schema_disruption]{Schema Disruption} nonetheless is present.

\begin{tcolorbox}[
    colback=white, 
    colframe=black, 
    sharp corners, 
    boxrule=0.5pt, 
    fonttitle=\bfseries
]
\begin{definition}[Schema Disruption]
\label{def:schema_disruption}
\textit{\textbf{Schema Disruption} is the phenomenon in which a language model is misled by the given examples due to a disrupted activation of the schema caused by the poor quality of the examples the model conditions on.}
\end{definition}
\end{tcolorbox}


\subsection{Schemas can be an effective method for knowledge retrieval}
Schemas also serve as a metric for retrieving relevant knowledge. \Cref{fig:end2end} showed an improvement in two of three datasets. The results illustrate the potential of using schema for in-context retrieval, along with the schema activation processes.

\ifthenelse{\equal{\papertype}{preprint}}{
    \begin{figure}
      \centering
      \includegraphics[width=0.4\linewidth]{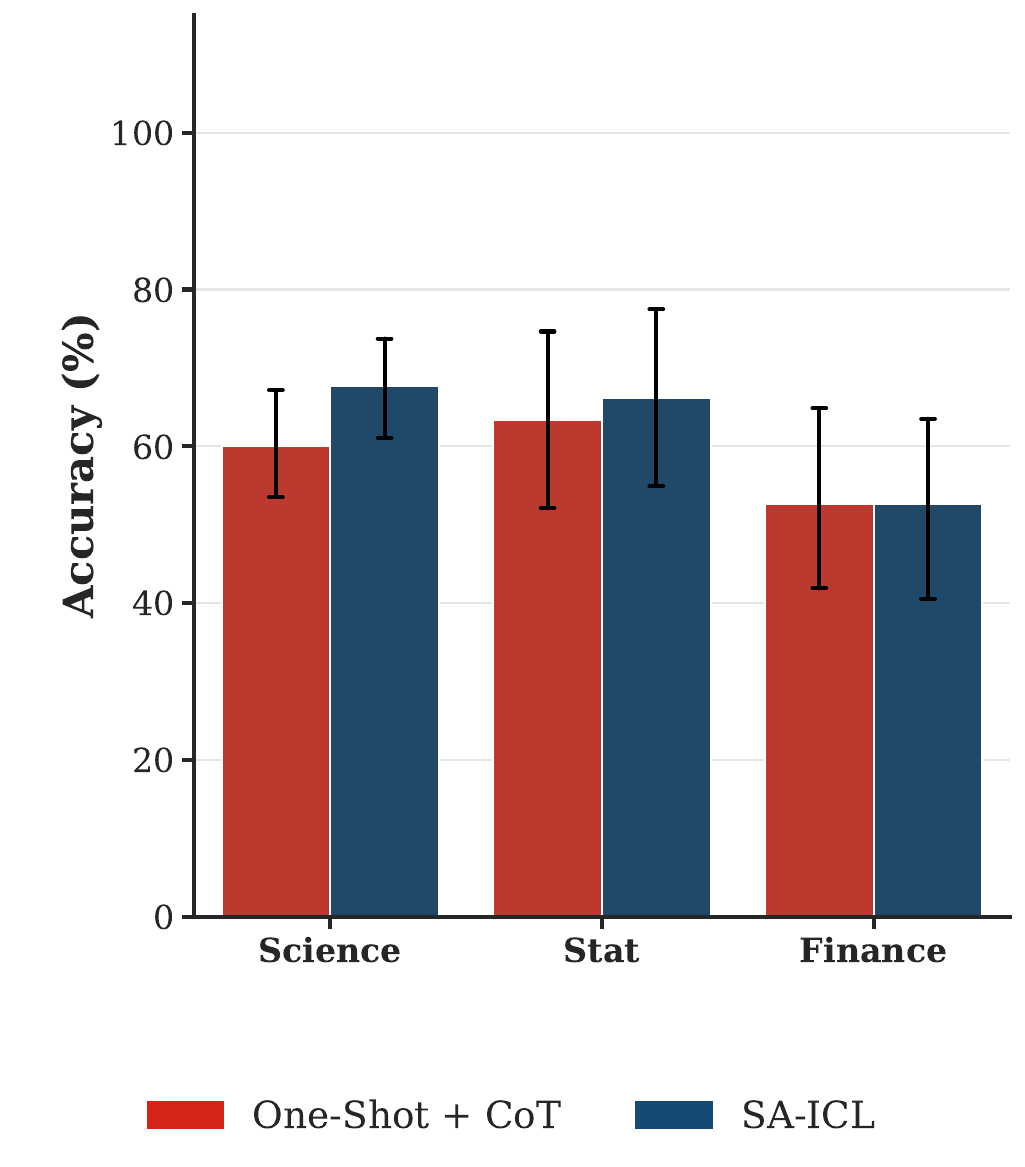}
      \caption{End-to-end performance of {\deepseek} utilizing {\cohere} for top-1 example retrieval via API calls.}
      \label{fig:end2end}
    \end{figure}
}
{
    \begin{figure}
      \centering
      \includegraphics[width=0.7\linewidth]{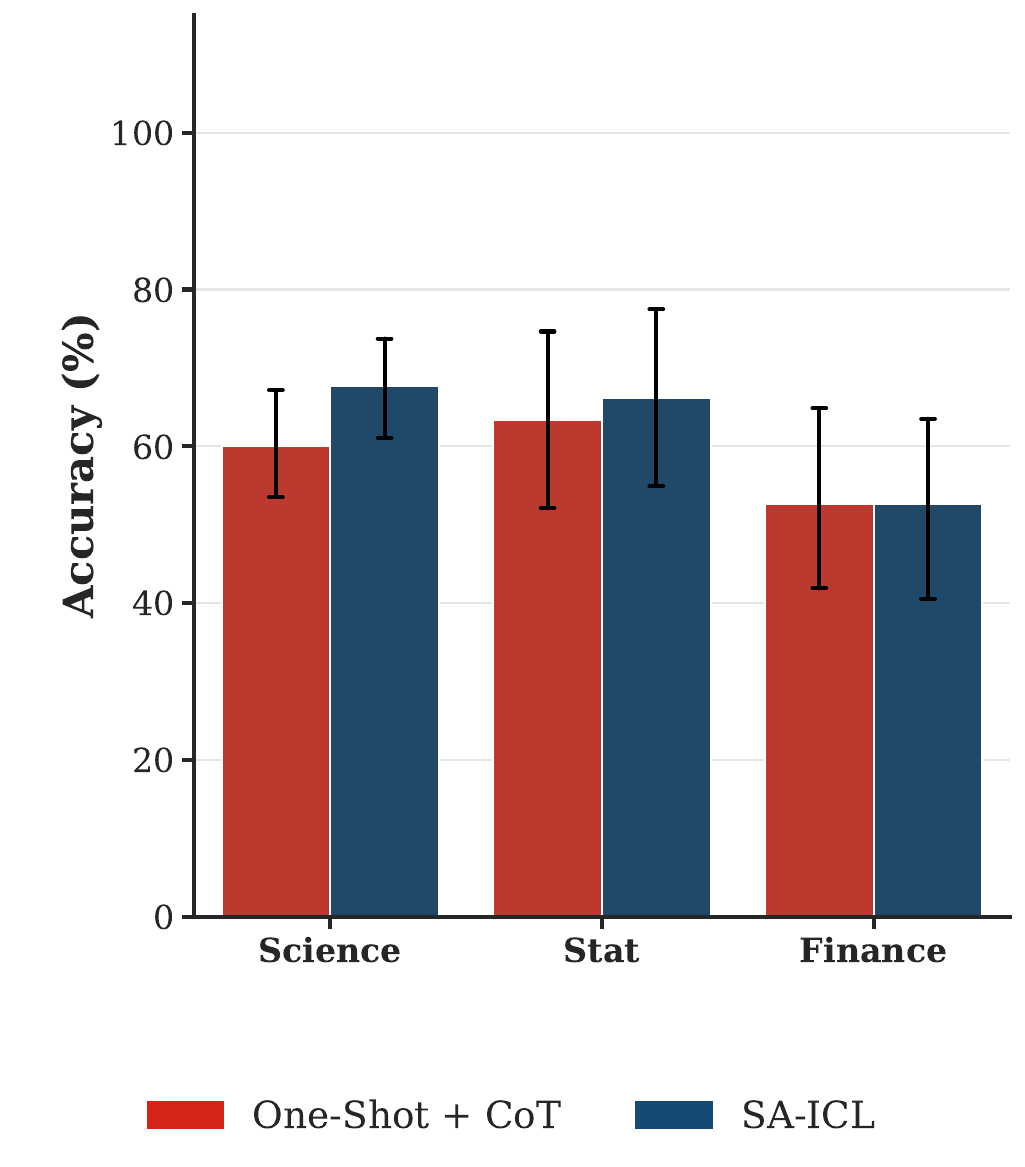}
      \caption{End-to-end performance of {\deepseek} utilizing {\cohere} for top-1 example retrieval via API calls.}
      \label{fig:end2end}
    \end{figure}
}

In another end-to-end scenario, we took a more aggressive approach by having LLM self-generate an example, a further step to simulate the human cognition process. We empirically observed that {\sysname} facilitates the use of self-generated examples more effectively (\Cref{fig:internal}).

\ifthenelse{\equal{\papertype}{preprint}}{
    \begin{figure}
      \centering
      \includegraphics[width=0.7\linewidth]{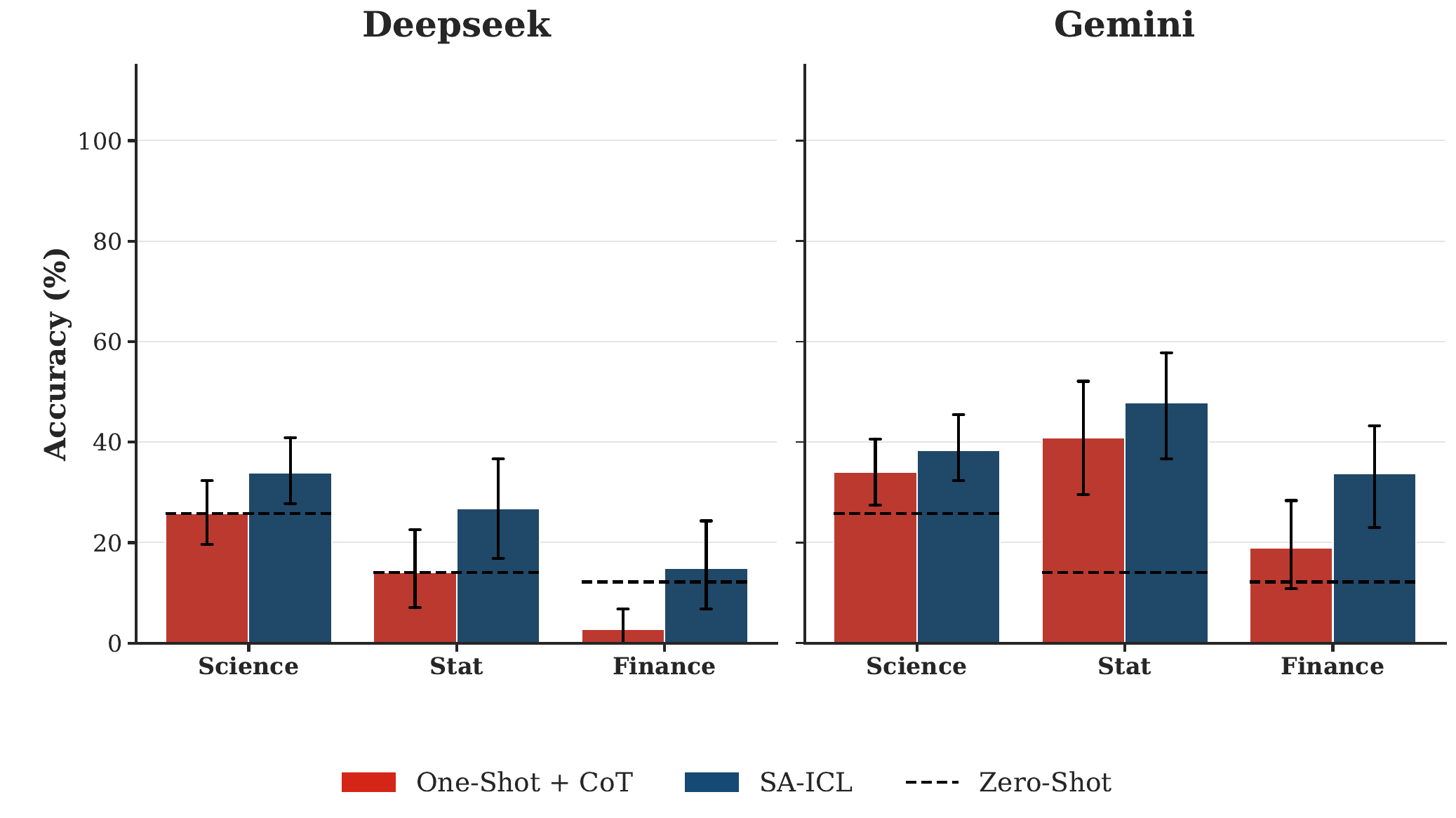}
      \caption{{\deepseek} and Gemini-2.0-Flash performance in an end-to-end scenario where the models self-generated internal examples first.}
      \label{fig:internal}
    \end{figure}
}
{
    \begin{figure}
      \centering
      \includegraphics[width=\linewidth]{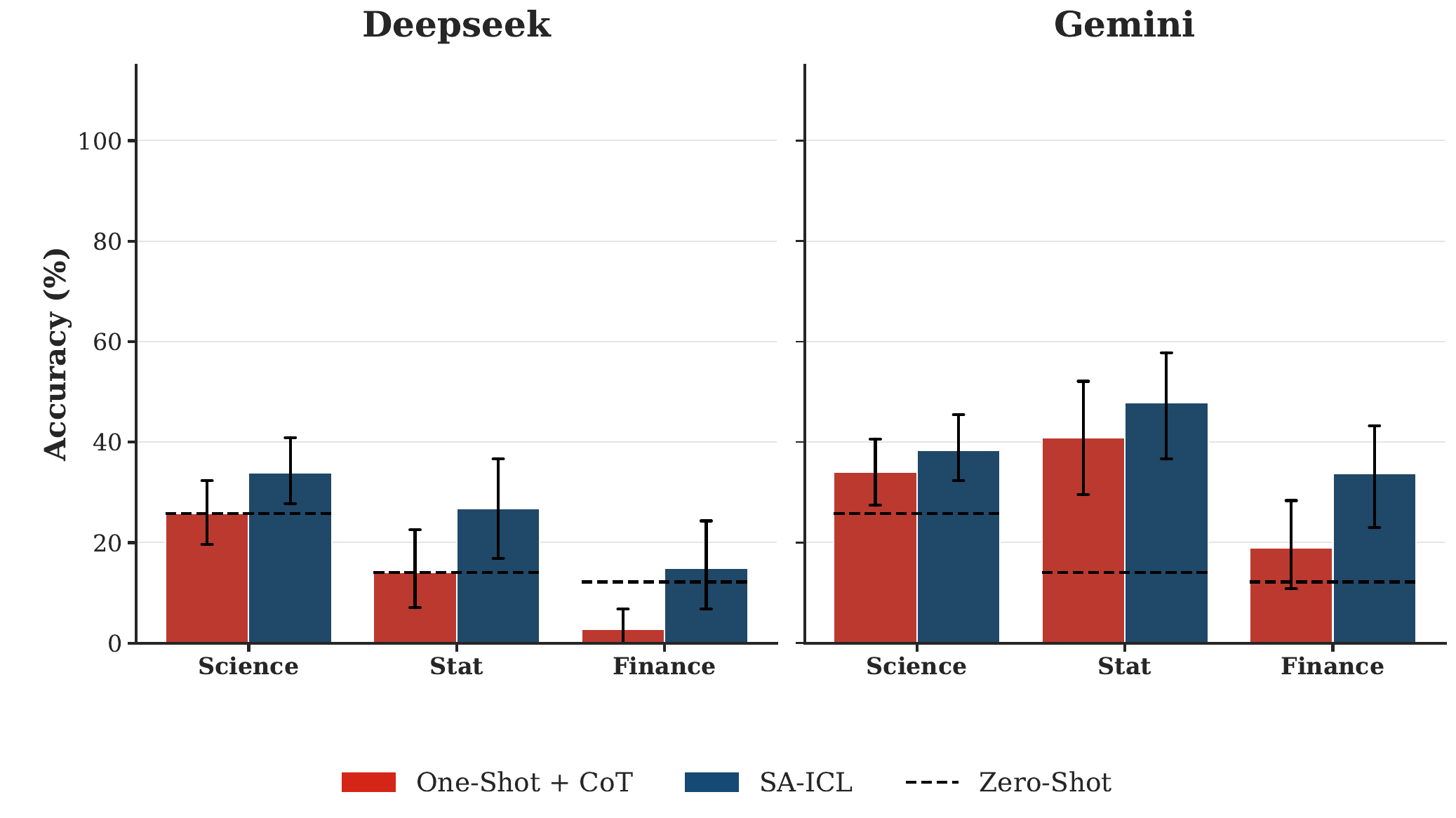}
      \caption{{\deepseek} and Gemini-2.0-Flash performance in an end-to-end scenario where the models self-generated internal examples first.}
      \label{fig:internal}
    \end{figure}
}



\subsection{Semantic drift and Schema Shift}
Finally, we quantify the semantic drift — or \textit{Schema Shift} — between latent and explicitly activated schemas in \Cref{app:schema_shift}. We observe that greater semantic differences strongly correlate with a higher probability of repairing incorrect reasoning paths, achieving up to a 7\% repair rate.

%% file: latex/sections/Results/pvalue_per_dataset.tex
\ifthenelse{\equal{\papertype}{preprint}}{
    \begin{table}[H]
        \centering
        \caption{Performance Impact Analysis by Domain --- {\sysname} results in a consistent better usage of prior knowledge across domains.}
        \label{tab:p_value_per_dataset}
        \resizebox{0.562\linewidth}{!}{%
            \begin{tabular}{l c c c l}
                \toprule
                \textbf{Metric} & \textbf{Baseline} & \textbf{Final} & \textbf{Change ($\Delta$)} & \textbf{$p$-value} \\
                \midrule
                \multicolumn{5}{l}{\textbf{Science}} \\
                \hspace{3mm} Reasoning Impact & 45.5\% & 46.1\% & +0.6\% & 0.5804 (ns) \\
                \hspace{3mm} Schema Impact    & 46.1\% & 51.0\% & +4.9\% & < 0.0001 (***) \\
                \midrule
                \multicolumn{5}{l}{\textbf{Statistics}} \\
                \hspace{3mm} Reasoning Impact & 44.4\% & 44.7\% & +0.3\% & 0.8674 (ns) \\
                \hspace{3mm} Schema Impact    & 44.7\% & 51.1\% & +6.4\% & 0.0004 (***) \\
                \midrule
                \multicolumn{5}{l}{\textbf{Finance}} \\
                \hspace{3mm} Reasoning Impact & 36.7\% & 36.0\% & -0.8\% & 0.6547 (ns) \\
                \hspace{3mm} Schema Impact    & 36.0\% & 40.3\% & +4.3\% & 0.0133 (*) \\
                \bottomrule
            \end{tabular}%
        }
    \end{table}
}{
    \begin{table}[H]
        \centering
        \caption{Performance Impact Analysis by Domain - {\sysname} results in a consistent better usage of prior knowledge across domains.}
        \label{tab:p_value_per_dataset}
        \resizebox{\columnwidth}{!}{%
            \begin{tabular}{l c c c l}
                \toprule
                \textbf{Metric} & \textbf{Baseline} & \textbf{Final} & \textbf{Change ($\Delta$)} & \textbf{$p$-value} \\
                \midrule
                \multicolumn{5}{l}{\textbf{Science}} \\
                \hspace{3mm} Reasoning Impact & 45.5\% & 46.1\% & +0.6\% & 0.5804 (ns) \\
                \hspace{3mm} Schema Impact    & 46.1\% & 51.0\% & +4.9\% & < 0.0001 (***) \\
                \midrule
                \multicolumn{5}{l}{\textbf{Statistics}} \\
                \hspace{3mm} Reasoning Impact & 44.4\% & 44.7\% & +0.3\% & 0.8674 (ns) \\
                \hspace{3mm} Schema Impact    & 44.7\% & 51.1\% & +6.4\% & 0.0004 (***) \\
                \midrule
                \multicolumn{5}{l}{\textbf{Finance}} \\
                \hspace{3mm} Reasoning Impact & 36.7\% & 36.0\% & -0.8\% & 0.6547 (ns) \\
                \hspace{3mm} Schema Impact    & 36.0\% & 40.3\% & +4.3\% & 0.0133 (*) \\
                \bottomrule
            \end{tabular}%
        }
    \end{table}
}

%% file: latex/figure_tables/trend.tex
\ifthenelse{\equal{\papertype}{preprint}}{
    \begin{figure}[H]
      \centering
      \includegraphics[width=0.5\linewidth]{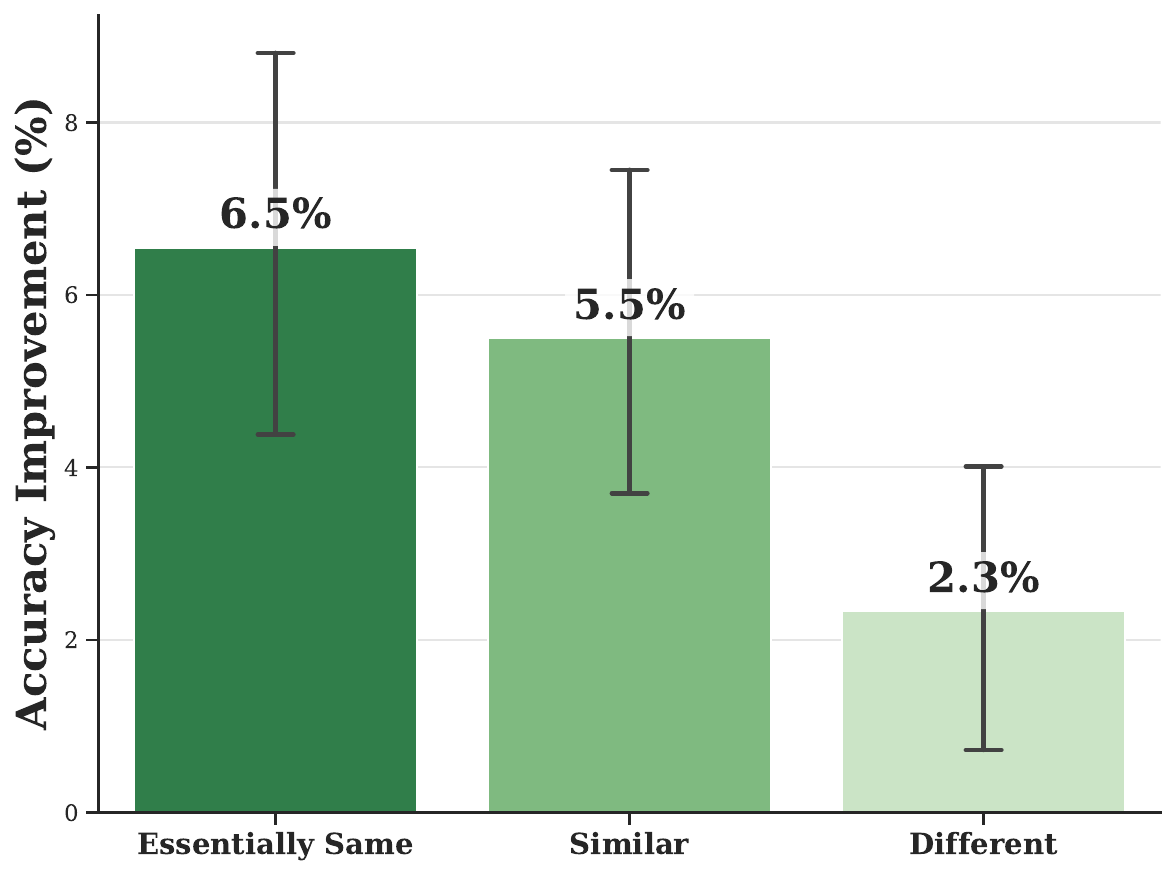}
      \caption{Aggregated improvements against {\OneShotCoT}, across the three datasets with all of the seven models tested. Surprisingly, when the example is of higher similarity, {\SchemaDormancy} becomes more obvious as a bottleneck, as the knowledge gap between the latent and activated schemas is highest (6.5\%).}
      \label{fig:trend}
    \end{figure}
}{
    \begin{figure}[H]
      \centering
      \includegraphics[width=\linewidth]{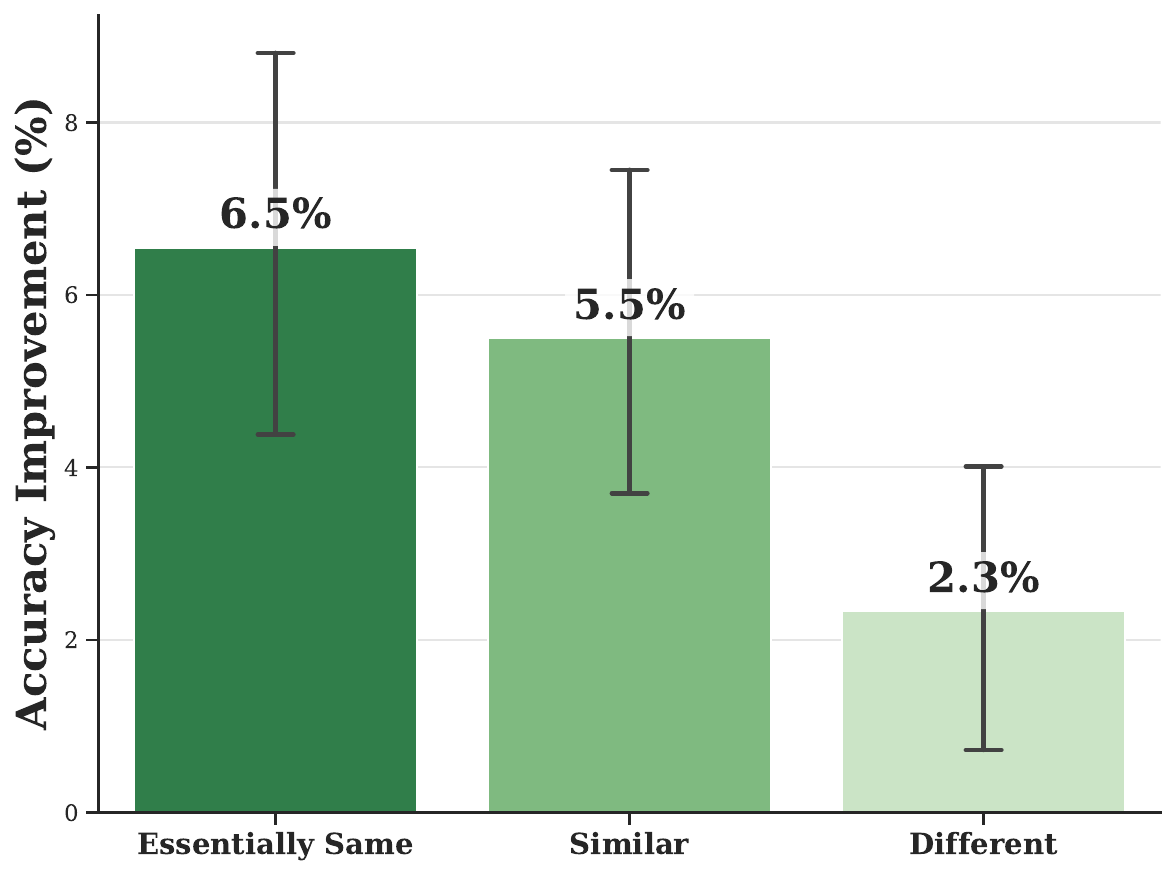}
      \caption{Aggregated improvements against {\OneShotCoT}, across the three datasets with all of the seven models tested. Surprisingly, when the example is of higher similarity, {\SchemaDormancy} becomes more obvious as a bottleneck, as the knowledge gap between the latent and activated schemas is highest (6.5\%).}
      \label{fig:trend}
    \end{figure}
}

%% file: latex/sections/Discussion_and_Future_work/discussion.tex
\section{Limitations \& Future work}


We encourage future work to explore agentic systems for dynamic schema generation so that {\sysname} can extend to more tailored schemas during inference time.

We also encourage future research to further validate {\sysname} on different types of tasks, such as long-form scientific explanation and the generation of mathematical proofs.


Finally, if future work can further verify that the internal behaviours of {\LLMs} are significantly affected by the schemas they activate and employ, much as in humans, {\sysname} could be useful for AI interpretability and safety.

\section{Conclusion}


{\sysname} bridges the gap between cognitive theory and the mechanistic realities of LLM reasoning. By replicating the {\WashingMachine} effect, we demonstrated that explicit schema activation intrinsically restructures attention dynamics, shifting models toward structural comprehension. Crucially, our comprehensive experiments validate {\sysname} as a unifying framework across diverse ICL methodologies. {\sysname} fundamentally redefines knowledge integration by moving from fragile instance-driven retrieval to robust abstraction-driven cognition.

Although beneficial in all scenarios, our analysis reveals that schema-based reasoning yields particularly strong gains for smaller models. We hypothesize that larger models are more rigidly anchored to their internal reasoning patterns — a phenomenon supported by our observations of Schema Disruption. Ultimately, these findings demonstrate that the next frontier of ICL should move beyond purely example-driven contexts to architectures that actively cultivate latent cognitive schemas.


%% file: latex/sections/acknowledgement.tex
A.A.-G. thanks Anders G. Fr{\o}seth for his generous support. A.A.-G. acknowledges the generous support of Natural Resources Canada and the Canada 150 Research Chairs program. P.C. acknowledges support from Google NSERC Industrial Research Chair Grants. The team appreciates the insights from Jingyu (Franklin) Hu, Qianfeng Wen, and Yifan Liu in the early stage of this project. This research was undertaken thanks in part to funding provided to the University of Toronto’s Acceleration Consortium from the Canada First Research Excellence Fund CFREF-2022-00042. Resources used in preparing this research were provided, in part, by the Province of Ontario, the Government of Canada through CIFAR, and companies sponsoring the Vector Institute.

%% file: latex/sections/appendix/appendix.tex
\renewcommand\thefigure{\thesection.\arabic{figure}}
\renewcommand\thetable{\thesection.\arabic{table}}

\setcounter{figure}{0}
\setcounter{table}{0}

\input{latex/sections/appendix/parts/preliminary}

\input{latex/sections/appendix/parts/interpretability}

\input{latex/sections/appendix/parts/math}

\input{latex/figure_tables/attention_entropy}
\input{latex/sections/appendix/parts/full_conversation}
\input{latex/sections/appendix/parts/exp_setup}

\input{latex/sections/appendix/parts/synthetic_data}
\input{latex/sections/appendix/parts/heatmap}

\input{latex/sections/appendix/parts/schema_template}

\input{latex/sections/appendix/parts/vision}
\input{latex/sections/appendix/parts/full_exp_analysis}

\input{latex/sections/appendix/parts/history}

\input{latex/sections/appendix/parts/semantic_shift}

\input{latex/sections/appendix/parts/aiuse}

%% file: latex/sections/appendix/parts/preliminary.tex
\clearpage
\section{Motivating and Preliminary experiments}
\subsection{{\WashingMachine} Experiment}
\label{app:washingmachine}
\begin{figure*}[t]
  \centering
  \includegraphics[width=\linewidth]{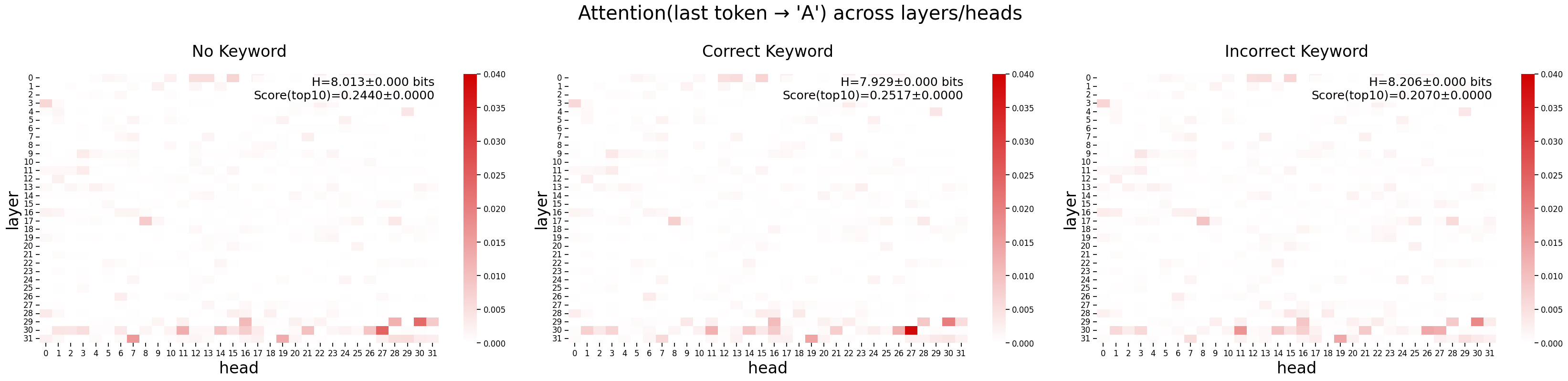}
  \includegraphics[width=\linewidth]{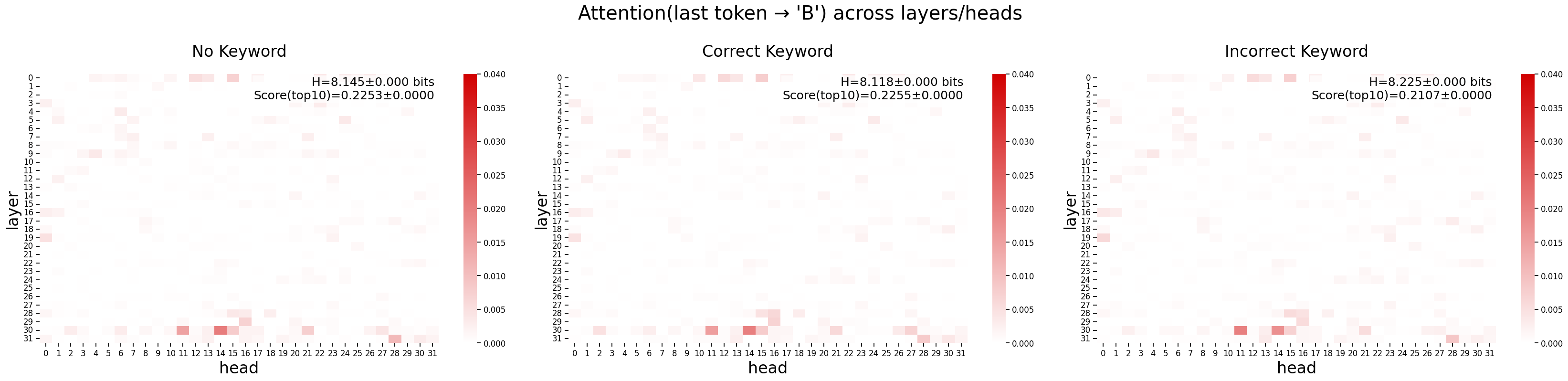}
  \caption{The full experiment results for the motivating experiment of the attention score heatmap. The correct keyword means including "Washing Machine" in the context.}
  \label{fig:interpretability_full}
\end{figure*}

Below is the full text of the ambiguous passage used in the classic “Washing Clothes” experiment in cognitive psychology and schema theory \citep{bransford1972contextual}:

\begin{quote}
\textit{The procedure is actually quite simple. First you arrange things into different groups. Of course, one pile may be sufficient depending on how much there is to do. If you have to go somewhere else due to lack of facilities that is the next step; otherwise you are pretty well set. It is important not to overdo things. That is, it is better to do too few things at once than too many. In the short run this may not seem important but complications can easily arise. A mistake can also be expensive. At first the whole procedure will seem complicated. Soon, however, it will become just another facet of life. It is difficult to foresee any end to the necessity for this task in the immediate future, but then one can never tell. After the procedure is completed one arranges the materials into different groups again. Then they can be put into their appropriate places. Eventually they will be used once more and the whole cycle will then have to be repeated. However, that is part of life.}
\end{quote}
We report all the attention heatmaps generated from our preliminary experiment in \ref{fig:interpretability_full}. The prompts used to generate the heatmaps on positive and negative label tokens in our preliminary experiment are also included below. 
\subsubsection{No Label}
\begin{lstlisting}
Answer with only one letter: A.Laundry and B.Dishwashing, choose one.

The procedure is actually quite simple. First you arrange items into different groups. Of course,one pile may be sufficient depending on how much there is to do. If you have to go somewhereelse due to lack of facilities that is the next step, otherwise, you are pretty well set. It is important not to overdo things. That is, it is better to do too few things at once than too many. In the short run this may not seem important, but complications can easily arise. A mistake can be expensive as well. At first, the whole procedure will seem complicated. Soon, however, it will become just another facet of life. It is difficult to foresee any end to the necessity for this task in the immediate future, but then, one never can tell. After the procedure is completed one arranges the materials into different groups again. Then they can be put into their appropriate places. Eventually they will be used once more and the whole cycle will then have to be repeated. However, this is part of life.
\end{lstlisting}
\subsubsection{Correct Label}
\begin{lstlisting}
Answer with only one letter: A.Laundry and B.Dishwashing, choose one.

Washing Machine. The procedure is actually quite simple. First you arrange items into different groups. Of course,one pile may be sufficient depending on how much there is to do. If you have to go somewhereelse due to lack of facilities that is the next step, otherwise, you are pretty well set. It is important not to overdo things. That is, it is better to do too few things at once than too many. In the short run this may not seem important, but complications can easily arise. A mistake can be expensive as well. At first, the whole procedure will seem complicated. Soon, however, it will become just another facet of life. It is difficult to foresee any end to the necessity for this task in the immediate future, but then, one never can tell. After the procedure is completed one arranges the materials into different groups again. Then they can be put into their appropriate places. Eventually they will be used once more and the whole cycle will then have to be repeated. However, this is part of life.
\end{lstlisting}
\subsubsection{Incorrect Label}
\begin{lstlisting}
Answer with only one letter: A.Laundry and B.Dishwashing, choose one.

Kitchen. The procedure is actually quite simple. First you arrange items into different groups. Of course,one pile may be sufficient depending on how much there is to do. If you have to go somewhereelse due to lack of facilities that is the next step, otherwise, you are pretty well set. It is important not to overdo things. That is, it is better to do too few things at once than too many. In the short run this may not seem important, but complications can easily arise. A mistake can be expensive as well. At first, the whole procedure will seem complicated. Soon, however, it will become just another facet of life. It is difficult to foresee any end to the necessity for this task in the immediate future, but then, one never can tell. After the procedure is completed one arranges the materials into different groups again. Then they can be put into their appropriate places. Eventually they will be used once more and the whole cycle will then have to be repeated. However, this is part of life.
\end{lstlisting}

\subsection{Preliminary results on GPQA-Chemistry and GPQA-Physics}
\label{app:preliminary}

As a first attempt to see whether there exists a knowledge gap when {\LLMs} were only given an example with no other information ({\OneShot}), we conducted preliminary experiments on GPQA-Chemistry and GPQA-Physics datasets. Unlike the experiments reported in the main text, we allowed the {\LLMs} to explicitly generate reasoning as part of their output. Also note that we were using Gemini 1.5-Flash, which has since been deprecated, for part of this preliminary experiment.

\input{latex/sections/Results/full_exp_tables}

%% file: latex/sections/Results/full_exp_tables.tex
\ifthenelse{\equal{\papertype}{preprint}}{
    \begin{table}[h]
    
    \label{tab:full_results}
    
    \centering
    
    \normalsize
    
    \setlength{\tabcolsep}{3pt} 
    
    \renewcommand{\arraystretch}{0.9} 
    
    \caption{Performances of {\sysname} and {\oneshot} on Chemistry and Physics questions. For Chemistry, the improvement is up to \textbf{39.67\%}, \textbf{34.88\%} for Physics. On average, the improvement in Chemistry is \textbf{9.81\%}, \textbf{12.91\%} for Physics. All values in the table were rounded up to the third decimal. For synthetic similarity experiments, we used {\oneshot} as the baseline comparison against {\sysname}. For latent similarity experiments, we used 1-shot, 3-shots, and 5-shots for the high, medium, and low similarity experiments, respectively, as the comparison against {\sysname}. Note that for Gemini, the One-Shot experiment in the Latent Similarity, we used Gemini 2.0 due to the Gemini 1.5 Flash being deprecated.}

    
    \resizebox{0.92\linewidth}{!}{%
    
    \begin{minipage}{\textwidth}

    
    \begin{subtable}{\textwidth}
    
    \centering
    
    \caption{Chemistry Results}
    
    \label{tab:chemistry-sub-comparison}
    
    \begin{tabular}{@{}ll|ccc|ccc@{}}
    
    \toprule
    
    \multirow{2}{*}{Model} & \multirow{2}{*}{Method} & \multicolumn{3}{c|}{Synthetic Similarity} & \multicolumn{3}{c}{Latent Similarity} \\
    
     &  & {\same} & {\similar} & {\different} & High & Medium & Low \\
    
    \midrule
    
    \multirow{2}{*}{Qwen-3} & {\sysname} & \textbf{0.763}  & \textbf{0.376}  & 0.301 & \textbf{0.667}  & \textbf{0.634} & 0.624 \\
    
     & {\oneshot} & 0.581  & 0.301  & 0.301 & 0.688  & 0.581 & 0.624 \\
    
    \midrule
    
    \multirow{2}{*}{Llama-3.1} & {\sysname} & \textbf{0.892}  & \textbf{0.430}  & \textbf{0.387} & \textbf{0.753}  & \textbf{0.548}  & \textbf{0.495} \\
    
     & {\oneshot} & 0.495  & 0.366  & 0.366 & 0.441  & 0.441  & 0.473 \\
    
    \midrule
    
    \multirow{2}{*}{GPT-4o Mini} & {\sysname} & \textbf{0.946}  & \textbf{0.462}  & \textbf{0.366} & \textbf{0.839}  & 0.581  & 0.559 \\
    
     & {\oneshot} & 0.688  & 0.366  & 0.323 & 0.667  & \textbf{0.624}  & \textbf{0.613} \\
    
    \midrule
    
    \multirow{2}{*}{GPT-4o} & {\sysname} & \textbf{0.925}  & 0.516  & \textbf{0.419} & \textbf{0.774}  & 0.581  & 0.667 \\
    
     & {\oneshot} & 0.656  & \textbf{0.559}  & 0.409 & 0.667  & \textbf{0.688}  & \textbf{0.699} \\
    
    \midrule
    
    \multirow{2}{*}{Ministral} & {\sysname} & \textbf{0.882}  & \textbf{0.473}  & \textbf{0.376} & \textbf{0.785}  & 0.634  & 0.624 \\
    
     & {\oneshot} & 0.731  & 0.376  & 0.280 & 0.677  & \textbf{0.656}  & \textbf{0.656} \\
    
    \midrule
    
    \multirow{2}{*}{Gemini 1.5 Flash$^*$} & {\sysname} & \textbf{0.796}  & \textbf{0.333}  & \textbf{0.280} & 0.452  & \textbf{0.663}  & \textbf{0.640} \\
    
     & {\oneshot} & 0.538  & 0.258  & 0.194 & \textbf{0.473}  & 0.452  & 0.409 \\
    
    \bottomrule
    
    \end{tabular}
    
    \end{subtable}

    \vspace{0.5em}

    
    \begin{subtable}{\textwidth}
    
    \centering
    
    \caption{Physics Results}
    
    \label{tab:physics-sub-comparison}
    
    \begin{tabular}{@{}ll|ccc|ccc@{}}
    
    \toprule
    
    \multirow{2}{*}{Model} & \multirow{2}{*}{Method} & \multicolumn{3}{c|}{Synthetic Similarity} & \multicolumn{3}{c}{Latent Similarity} \\
    
     &  & {\same} & {\similar} & {\different} & High & Medium & Low \\
    
    \midrule
    
    \multirow{2}{*}{Qwen-3} & {\sysname} & \textbf{0.756}  & 0.465  & 0.349 & \textbf{0.686}  & \textbf{0.814}  & \textbf{0.721} \\
    
     & {\oneshot} & 0.616  & \textbf{0.477}  & \textbf{0.477} & 0.605  & 0.581  & 0.581 \\
    
    \midrule
    
    \multirow{2}{*}{Llama-3.1} & {\sysname} & \textbf{0.814}  & \textbf{0.430}  & \textbf{0.407} & \textbf{0.744}  & \textbf{0.535}  & \textbf{0.605} \\
    
     & {\oneshot} & 0.512  & 0.314  & 0.372 & 0.465  & 0.395  & 0.547 \\
    
    \midrule
    
    \multirow{2}{*}{GPT-4o Mini} & {\sysname} & \textbf{0.977}  & \textbf{0.512}  & \textbf{0.523} & \textbf{0.872}  & 0.581  & \textbf{0.628} \\
    
     & {\oneshot} & 0.628  & 0.372  & 0.372 & 0.593  & \textbf{0.593}  & 0.547 \\
    
    \midrule
    
    \multirow{2}{*}{GPT-4o} & {\sysname} & \textbf{0.953}  & \textbf{0.663}  & \textbf{0.616} & \textbf{0.872}  & 0.674  & 0.698 \\
    
     & {\oneshot} & 0.721  & 0.616  & 0.547 & 0.744  & \textbf{0.698}  & \textbf{0.721} \\
    
    \midrule
    
    \multirow{2}{*}{Ministral} & {\sysname} & \textbf{0.930}  & \textbf{0.535}  & \textbf{0.488} & \textbf{0.849}  & 0.686  & 0.624 \\
    
     & {\oneshot} & 0.802  & 0.372  & 0.256 & 0.663  & \textbf{0.721}  & \textbf{0.686} \\
    
    \midrule
    
    \multirow{2}{*}{Gemini 1.5 Flash$^*$} & {\sysname} & \textbf{0.767}  & \textbf{0.407}  & \textbf{0.360} & \textbf{0.655}  & \textbf{0.559}  & \textbf{0.559} \\
    
     & {\oneshot} & 0.500  & 0.349  & 0.349 & 0.500  & 0.488  & 0.512 \\
    
    \bottomrule
    
    \end{tabular}
    
    \end{subtable}

    \end{minipage}%
    
    } 
    
    \end{table}
}{
\begin{table}[h]
    
    \label{tab:full_results}
    
    \centering
    
    \normalsize
    
    \setlength{\tabcolsep}{3pt} 
    
    \renewcommand{\arraystretch}{0.9} 
    
    \caption{Performances of {\sysname} and {\oneshot} on Chemistry and Physics questions. For Chemistry, the improvement is up to \textbf{39.67\%}, \textbf{34.88\%} for Physics. On average, the improvement in Chemistry is \textbf{9.81\%}, \textbf{12.91\%} for Physics. All values in the table were rounded up to the third decimal. For synthetic similarity experiments, we used {\oneshot} as the baseline comparison against {\sysname}. For latent similarity experiments, we used 1-shot, 3-shots, and 5-shots for the high, medium, and low similarity experiments, respectively, as the comparison against {\sysname}. Note that for Gemini, the One-Shot experiment in the Latent Similarity, we used Gemini 2.0 due to the Gemini 1.5 Flash being deprecated.}

    
    \resizebox{\columnwidth}{!}{%
    
    \begin{minipage}{\textwidth}

    
    \begin{subtable}{\textwidth}
    
    \centering
    
    \caption{Chemistry Results}
    
    \label{tab:chemistry-sub-comparison}
    
    \begin{tabular}{@{}ll|ccc|ccc@{}}
    
    \toprule
    
    \multirow{2}{*}{Model} & \multirow{2}{*}{Method} & \multicolumn{3}{c|}{Synthetic Similarity} & \multicolumn{3}{c}{Latent Similarity} \\
    
     &  & {\same} & {\similar} & {\different} & High & Medium & Low \\
    
    \midrule
    
    \multirow{2}{*}{Qwen-3} & {\sysname} & \textbf{0.763}  & \textbf{0.376}  & 0.301 & \textbf{0.667}  & \textbf{0.634} & 0.624 \\
    
     & {\oneshot} & 0.581  & 0.301  & 0.301 & 0.688  & 0.581 & 0.624 \\
    
    \midrule
    
    \multirow{2}{*}{Llama-3.1} & {\sysname} & \textbf{0.892}  & \textbf{0.430}  & \textbf{0.387} & \textbf{0.753}  & \textbf{0.548}  & \textbf{0.495} \\
    
     & {\oneshot} & 0.495  & 0.366  & 0.366 & 0.441  & 0.441  & 0.473 \\
    
    \midrule
    
    \multirow{2}{*}{GPT-4o Mini} & {\sysname} & \textbf{0.946}  & \textbf{0.462}  & \textbf{0.366} & \textbf{0.839}  & 0.581  & 0.559 \\
    
     & {\oneshot} & 0.688  & 0.366  & 0.323 & 0.667  & \textbf{0.624}  & \textbf{0.613} \\
    
    \midrule
    
    \multirow{2}{*}{GPT-4o} & {\sysname} & \textbf{0.925}  & 0.516  & \textbf{0.419} & \textbf{0.774}  & 0.581  & 0.667 \\
    
     & {\oneshot} & 0.656  & \textbf{0.559}  & 0.409 & 0.667  & \textbf{0.688}  & \textbf{0.699} \\
    
    \midrule
    
    \multirow{2}{*}{Ministral} & {\sysname} & \textbf{0.882}  & \textbf{0.473}  & \textbf{0.376} & \textbf{0.785}  & 0.634  & 0.624 \\
    
     & {\oneshot} & 0.731  & 0.376  & 0.280 & 0.677  & \textbf{0.656}  & \textbf{0.656} \\
    
    \midrule
    
    \multirow{2}{*}{Gemini 1.5 Flash^{*}} & {\sysname} & \textbf{0.796}  & \textbf{0.333}  & \textbf{0.280} & 0.452  & \textbf{0.663}  & \textbf{0.640} \\
    
     & {\oneshot} & 0.538  & 0.258  & 0.194 & \textbf{0.473}  & 0.452  & 0.409 \\
    
    \bottomrule
    
    \end{tabular}
    
    \end{subtable}

    \vspace{0.5em}

    
    \begin{subtable}{\textwidth}
    
    \centering
    
    \caption{Physics Results}
    
    \label{tab:physics-sub-comparison}
    
    \begin{tabular}{@{}ll|ccc|ccc@{}}
    
    \toprule
    
    \multirow{2}{*}{Model} & \multirow{2}{*}{Method} & \multicolumn{3}{c|}{Synthetic Similarity} & \multicolumn{3}{c}{Latent Similarity} \\
    
     &  & {\same} & {\similar} & {\different} & High & Medium & Low \\
    
    \midrule
    
    \multirow{2}{*}{Qwen-3} & {\sysname} & \textbf{0.756}  & 0.465  & 0.349 & \textbf{0.686}  & \textbf{0.814}  & \textbf{0.721} \\
    
     & {\oneshot} & 0.616  & \textbf{0.477}  & \textbf{0.477} & 0.605  & 0.581  & 0.581 \\
    
    \midrule
    
    \multirow{2}{*}{Llama-3.1} & {\sysname} & \textbf{0.814}  & \textbf{0.430}  & \textbf{0.407} & \textbf{0.744}  & \textbf{0.535}  & \textbf{0.605} \\
    
     & {\oneshot} & 0.512  & 0.314  & 0.372 & 0.465  & 0.395  & 0.547 \\
    
    \midrule
    
    \multirow{2}{*}{GPT-4o Mini} & {\sysname} & \textbf{0.977}  & \textbf{0.512}  & \textbf{0.523} & \textbf{0.872}  & 0.581  & \textbf{0.628} \\
    
     & {\oneshot} & 0.628  & 0.372  & 0.372 & 0.593  & \textbf{0.593}  & 0.547 \\
    
    \midrule
    
    \multirow{2}{*}{GPT-4o} & {\sysname} & \textbf{0.953}  & \textbf{0.663}  & \textbf{0.616} & \textbf{0.872}  & 0.674  & 0.698 \\
    
     & {\oneshot} & 0.721  & 0.616  & 0.547 & 0.744  & \textbf{0.698}  & \textbf{0.721} \\
    
    \midrule
    
    \multirow{2}{*}{Ministral} & {\sysname} & \textbf{0.930}  & \textbf{0.535}  & \textbf{0.488} & \textbf{0.849}  & 0.686  & 0.624 \\
    
     & {\oneshot} & 0.802  & 0.372  & 0.256 & 0.663  & \textbf{0.721}  & \textbf{0.686} \\
    
    \midrule
    
    \multirow{2}{*}{Gemini 1.5 Flash^{*}} & {\sysname} & \textbf{0.767}  & \textbf{0.407}  & \textbf{0.360} & \textbf{0.655}  & \textbf{0.559}  & \textbf{0.559} \\
    
     & {\oneshot} & 0.500  & 0.349  & 0.349 & 0.500  & 0.488  & 0.512 \\
    
    \bottomrule
    
    \end{tabular}
    
    \end{subtable}

    \end{minipage}%
    
    } 
    
    \end{table}
}

%% file: latex/sections/appendix/parts/interpretability.tex
\section{Interpretability Experiments}
\label{app:interpretability_exps}
\subsection{Report on token  counts and correctness for the first 10 {\chemistrydataset} questions}

The interpretability experiments are conducted via logit-centric methods to get the overall confidence of LLM generation by retrieving the token-level log likelihood \citep{Zhang2025TokenLevelUE} from the OpenAI chat completions API for each of our tested ICL prompting strategies. We implemented a customized chat interface to display the log-probability in coloured texts using the colour schema provided in the legend in \Cref{fig:interpretability} for better visualization and interaction.

From \Cref{tab:token-counts}, we observed that {\sysname} achieves the highest accuracy in these questions while simultaneously generating fewer final step output tokens compared to {\OneShotCoT}, except for the sixth question.

By increasing internal iteration depth, we maximize the signal-to-noise ratio of the assistant's final response to the user in order to minimize the cognitive load on the user. Through {\sysname}, we use the intermediate black box to ensure logic is implicit in a correct, brief answer rather than explicit in a verbose trace.

\subsection{Visualization of the log-probabilities of each token in the solver's output}

To investigate the effectiveness of schema-based learning over CoT and {\oneshot} prompting methods from an interpretability standpoint, we logged the log likelihood of the top-5 predicted tokens from the LLM generation output in \Cref{fig:interpretability} \citep{Zhang2025TokenLevelUE}. {\oneshot} prompting often encourages the model to overfit to the given demonstration by focusing on fitting the output format (e.g., Answer: {ANSWER}), rather than leveraging the example to activate prior knowledge and understandings to reason the problem better and provide well-thought-out answers. This problem is discussed in earlier work as pattern-matching. 

While prior work noted that enforcing rigid, structured outputs can reduce the flexibility needed for effective reasoning, {\sysname} reached equilibrium between structured reasoning and free-form thinking. \Cref{tab:token-counts} showed that in high-knowledge settings, where the answer is already well represented in the context, CoT may introduce unnecessary verbosity and even hurt performance, whereas {\sysname} provides more direct and efficient knowledge activation.

\input{latex/sections/Results/interpretability_table}
\ifthenelse{\equal{\papertype}{preprint}}{
    \begin{figure}[H]
      \centering
      \includegraphics[width=0.8\linewidth]{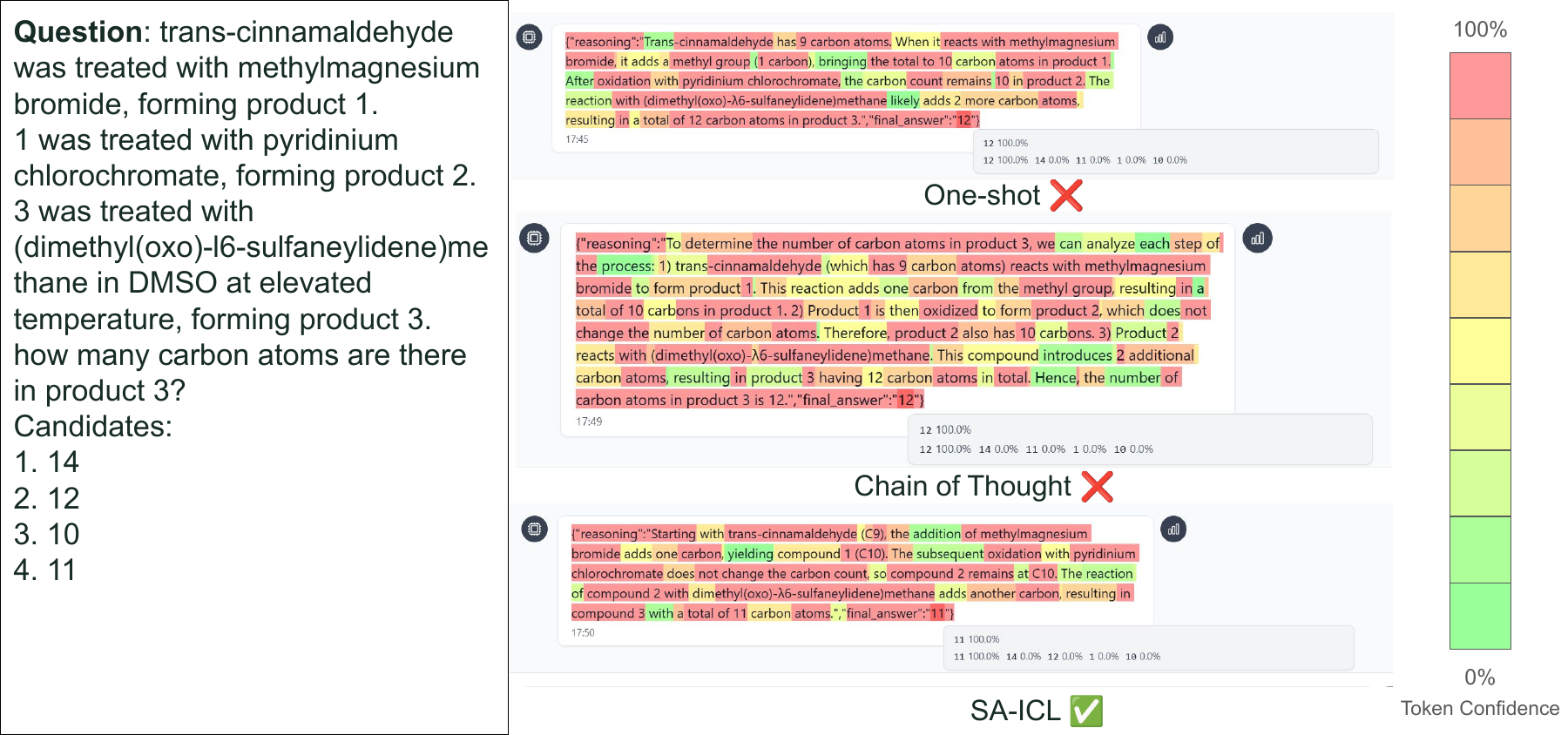}
      \caption{The probabilities of each token in the problem solver output. Note that only {\sysname} confidently gave the correct answer for this question, while pure {\oneshot} or {\oneshot} + CoT confidently gave the incorrect answer.}
      \label{fig:interpretability}
    \end{figure}
}{
    \begin{figure}[H]
      \centering
      \includegraphics[width=\linewidth]{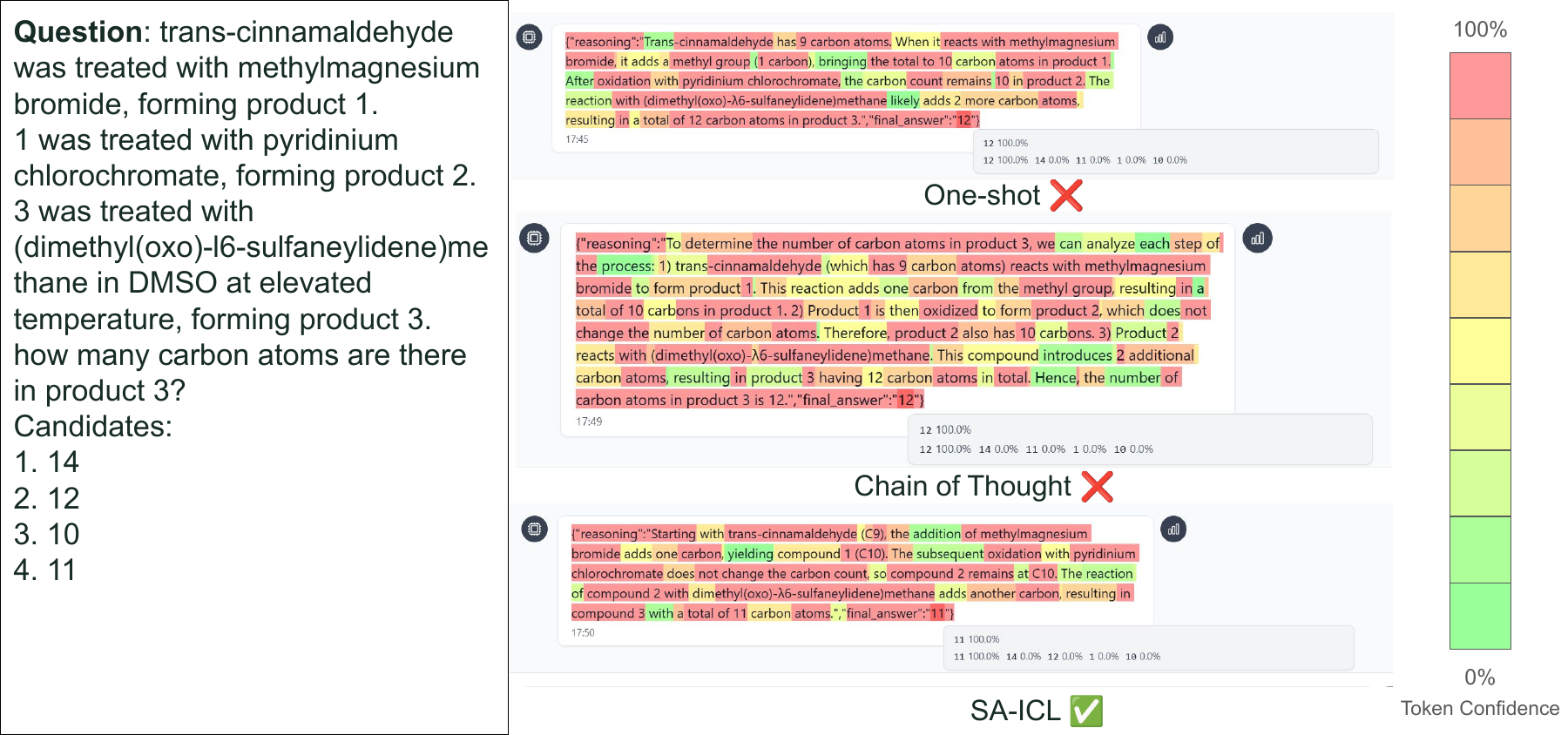}
      \caption{The probabilities of each token in the problem solver output. Note that only {\sysname} confidently gave the correct answer for this question, while pure {\oneshot} or {\oneshot} + CoT confidently gave the incorrect answer.}
      \label{fig:interpretability}
    \end{figure}
}

%% file: latex/sections/Results/interpretability_table.tex
\begin{table*}[h!]
\small
\centering
\caption{Token counts and correctness (\cmark/\xmark) across different prompting strategies for the first 10 questions from {\chemistrydataset} dataset using GPT-4o Mini with temperature set to 0 for the most consistent results. All prior knowledge in this experiment is retrieved using {\high} in Latent Similarity. We ran \textbf{three} experiments per question to get the average token counts, and we used the majority correctness as overall correctness.}
\label{tab:token-counts}
\begin{tabular}{>{\raggedright\arraybackslash}p{4cm}cccccc}
\toprule
\textbf{Question ID} 
& \multicolumn{2}{c}{\textbf{{\sysname}}} 
& \multicolumn{2}{c}{\textbf{{\oneshot}}} 
& \multicolumn{2}{c}{\textbf{{\oneshot} + CoT}} \\
\cmidrule(lr){2-3} \cmidrule(lr){4-5} \cmidrule(lr){6-7}
& \textbf{Tokens} & \textbf{Correct} 
& \textbf{Tokens} & \textbf{Correct} 
& \textbf{Tokens} & \textbf{Correct} \\
\midrule

2662eff7a6231613f...caaeb & 150 & \cmark  & 133 & \xmark & 196 & \xmark \\
fc081c2fbb63be500...65420 & 161 & \cmark  & 132 & \cmark  & 206 & \cmark  \\
a8be7a4963bfb6bc7...99122 & 180 & \cmark  & 156 & \cmark  & 228 & \cmark  \\
f730b35adb897658b...a77e5 & 166 & \cmark  & 412 & \xmark & 417 & \xmark \\
1ce3d847d25b2c2f6...01155 & 231 & \cmark  & 235 & \cmark  & 254 & \cmark  \\
d8c36bd55ba561cb4...7a049 & 308 & \cmark  & 324 & \cmark  & 273 & \cmark  \\
40b2b50a3c993902d...0bfcb &  91 & \cmark  & 138 & \cmark  & 186 & \cmark  \\
a2136b05b78259562...184d7 &  76 & \cmark  &  76 & \xmark &  91 & \xmark \\
cbf5c336a0990294b...7d447 & 203 & \cmark  & 170 & \xmark  & 222 & \xmark \\
16464cac7090a24d3...9bafd2 & 175 & \cmark  & 154 & \cmark  & 217 & \cmark  \\
\midrule
\textbf{Total Correct} & \multicolumn{2}{c}{\textbf{10/10}} & \multicolumn{2}{c}{\textbf{6/10}} & \multicolumn{2}{c}{\textbf{6/10}} \\
\bottomrule
\end{tabular}
\end{table*}

%% file: latex/sections/appendix/parts/math.tex
\section{Mathematical Formalization of {\sysnamefull}}
\label{app:algorithm_walkthrough}
\subsection{Memory: Schemas and Episodic Traces}
We model memory as a bipartite structure linking abstract schemas to multiple episodic examples:
\[
\mathcal{M} = \{ (\mathcal{S}_i, e_j, w_{ij}(t)) \mid \mathcal{S}_i \in \mathcal{S}, \; e_j \in \mathcal{E} \},
\]
where
\begin{itemize}
    \item $\mathcal{S} = \{\mathcal{S}_1, \ldots, \mathcal{S}_N\}$ is the set of schemas,
    \item $\mathcal{E} = \{e_1, \ldots, e_M\}$ is the set of episodic traces (examples),
    \item $w_{ij}(t) \in [0,1]$ is the association strength between schema $\mathcal{S}_i$ and example $e_j$ at time $t$.
\end{itemize}
Association weights decay over time, modelling episodic forgetting:
\[
w_{ij}(t) = w_{ij}(0) \cdot \exp(-\lambda t), \quad \lambda > 0.
\]

The exponential function is an estimate for the forgetting curve, as we want the examples learned earlier to have less impact.

\subsection{Problem Representation (Schema Formation)}
Given an input problem $x$, the LLM constructs a mental representation (schema):
\[
\mathcal{S}_x = \mathcal{R}(x)
\]
where $\mathcal{R}$ is an embedding or representation function.

\subsection{Prior Schema Retrieval}
\label{app:prior_schema_retrieval}
The model retrieves a schema $\hat{\mathcal{S}}$ from $\mathcal{S}$:
\[
\hat{\mathcal{S}} = \arg\max_{\mathcal{S}_i\in \mathcal{S}} \, \mathrm{sim}(\mathcal{S}_x, \mathcal{S}_i),
\]
where $\mathrm{sim}(\cdot,\cdot)$ denotes the similarity function, which may be cosine similarity, re-ranking, or another metric.

\subsection{Episodic Examples Retrieval}
For notational convenience, let $\hat{\imath} \in \{1, ..., N\}$ denote the index of $\hat{\mathcal{S}}$ such that $\hat{\mathcal{S}}=\mathcal{S}_{\hat{\imath}}$. More formally,
\[
\hat{\imath} = \arg\max_{i \in \{1,\dots,N\}} \, \mathrm{sim}(\mathcal{S}_x, \mathcal{S}_i).
\]

Afterwards, given the retrieved schema $\hat{\mathcal{S}}=\mathcal{S}_{\hat{\imath}}$, {\sysname} selects all episodic examples whose (decayed) association to $\hat{\mathcal{S}}$ exceeds a threshold $\tau \in [0,1]$:
\[
\hat{\mathcal{E}}_{\tau}(t \mid \hat{\imath})
\;=\;
\big\{\, e_j \in \mathcal{E} \;:\; w_{\hat{\imath} j}(t) \ge \tau \,\big\}.
\]

\subsection{Schema Activation (Assimilation / Accommodation)}
The retrieved schema and its selected episodic set guide activation of a new schema for the current problem:
\[
\mathcal{S}_{\text{new}} = f\!\big(\mathcal{S}_x, \hat{\mathcal{S}}, \hat{\mathcal{E}}_{\tau}(t \mid \hat{\imath})\big),
\]
where $f$ denotes the integration mechanism.

Then, \textit{assimilation} and \textit{accommodation} can be conceptualized as follows:
\begin{itemize}
    \item \textbf{Assimilation:} $\mathcal{S}_{\text{new}} \approx \mathcal{S}_x$ when $\hat{\mathcal{S}}$ fits well.
    \item \textbf{Accommodation:} $\mathcal{S}_{\text{new}}$ requires restructuring when fit is poor.
\end{itemize}

\subsection{Schema-Guided Problem Solving}
Finally, the LLM produces an output conditioned on the activated schema:
\[
y = \mathrm{LLM}\!\big(x, \mathcal{S}_{\text{new}}\big).
\]

\subsection{End-to-End Equation}
Combining all steps of our framework (i.e., schema-first, then thresholded episodic selection), we obtain the following equation representing the full {\sysname} inference procedure:

\ifthenelse{\equal{\papertype}{preprint}}{
    \begin{equation}
    y = \mathrm{LLM}\Big( x,\; f\Big( \mathcal{R}(x),\; \\
    \underbrace{\arg\max_{\mathcal{S}_i\in \mathcal{S}}\mathrm{sim}(\mathcal{R}(x),\mathcal{S}_i)}_{\hat{\mathcal{S}}=\mathcal{S}_{\hat{\imath}}}, 
    \underbrace{\big\{\, e_j \in \mathcal{E} \;:\; w_{\hat{\imath} j}(t) \ge \tau \,\big\}}_{\hat{\mathcal{E}}_{\tau}(t \mid \hat{\imath})} \Big) \Big).
    \end{equation}
}{
    \begin{multline}
    y = \mathrm{LLM}\Big( x,\; f\Big( \mathcal{R}(x),\; \\
    \underbrace{\arg\max_{\mathcal{S}_i\in \mathcal{S}}\mathrm{sim}(\mathcal{R}(x),\mathcal{S}_i)}_{\hat{\mathcal{S}}=\mathcal{S}_{\hat{\imath}}}, 
    \underbrace{\big\{\, e_j \in \mathcal{E} \;:\; w_{\hat{\imath} j}(t) \ge \tau \,\big\}}_{\hat{\mathcal{E}}_{\tau}(t \mid \hat{\imath})} \Big) \Big).
    \end{multline}
}

\subsection{Assumptions for this work}

We acknowledge that, throughout our investigations, we made several assumptions in the interest of experimental control, reproducibility, and variable isolation. Most notably, schema templates were manually selected rather than retrieved via the mechanism described in \hyperref[app:prior_schema_retrieval]{Section~C.3}. Furthermore, we evaluated four different templates in total and used CoT as the primary template for all main experiments. These design decisions allowed us to conduct systematic experiments, such as our ablation study isolating the effect of schema activation across templates, and the consistent improvements observed across seven diverse models and three domains suggest that our core findings are robust.

%% file: latex/sections/appendix/parts/full_conversation.tex
\section{A full conversation history}
\label{app:full_conversation_solver}
To ensure experimental reproducibility, the full logs will be uploaded to a GitHub repository before publication, where readers can find the prompts for the schema activations and the problem-solving.

Here, we included one specific example of {\sysname} at inference:
\begin{lstlisting}[language=Python]
"""
  "messages": [
    {
      "role": "system",
      "content": "You will be givin an example question along with its schema. Then, you will be given a new question with a preliminary schema. Think about if you could refine the preliminary schema based on the example question. Then, select the most appropriate answer for the new question."
    },
    {
      "role": "user",
      "content": [
        {
          "type": "text",
          "text": "Question: A proton is moving along the vertical axis upwards with a Lorentz factor of 3. An antiproton is moving along the vertical axis downwards with a Lorentz factor of 3. After colliding, both annihilate producing 2 photons.\n\nThe energies of both photons were measured and they have the same value. The directions were also measured. One of the photons was moving in the upper-left direction, while another in the upper-right direction.\n\nWhat is the angle between the directions of the photons?\nCandidates:\n1. 180 degrees\n2. 150 degrees\n3. 90 degrees\n4. 120 degrees\nAnswer to this question is: 120 degrees#### Schema:\n##### facts:\nProtons and antiprotons are fundamental particles with intrinsic properties such as charge, mass, and spin. A Lorentz factor is a concept from special relativity that accounts for the effects of high-speed motion as measured by observers in different reference frames. In particle collisions, energy and momentum conservation laws apply. When particles annihilate, they convert their mass into other forms of energy, such as photons. Photons are massless particles that travel at the speed of light and carry energy and momentum. The direction of a photon's movement can be described using vector components or angles relative to a reference frame.\n"
        },
        {
          "type": "text",
          "text": "\nBelow is the new question with a preliminary schema\n"
        },
        {
          "type": "text",
          "text": "Question: An electron is moving along the horizontal axis from the left with a Lorentz factor of 4. A positron is also moving along the horizontal axis, but from the right with a Lorentz factor of 2. After colliding, both annihilate producing 2 photons.\n\nThe energies of both photons were measured and they have the same value. The directions were also measured. One of the photons was moving in the upper-right direction, while another in the lower-right direction.\n\nWhat is the angle between the directions of the photons?\nCandidates:\n1. 74\n2. 96\n3. 172\n4. 138\n"
        },
        {
          "type": "text",
          "text": "Below is the preliminary schema:\n"
        },
        {
          "type": "text",
          "text": "#### Schema:\n##### facts:\nAn electron moving along a horizontal axis with a Lorentz factor of 4, A positron moving along the same axis from the right with a Lorentz factor of 2, Both annihilate into two photons, The energies of both photons are equal, One photon moves in an upper-right direction, another in a lower-right direction.\n"
        },
        {
          "type": "text",
          "text": "\nThink about if you could refine the preliminary schema based on the example provided above."
        }
      ]
    },
    {
      "role": "assistant",
      "content": "#### Schema:\n##### facts:\nElectrons and positrons are fundamental particles with intrinsic properties such as charge, mass, and spin. A Lorentz factor is a concept from special relativity that accounts for the effects of high-speed motion as measured by observers in different reference frames. In particle collisions, energy and momentum conservation laws apply. When particles annihilate, they convert their mass into other forms of energy, such as photons. Photons are massless particles that travel at the speed of light and carry energy and momentum. The direction of a photon's movement can be described using vector components or angles relative to a reference frame.\n"
    },
    {
      "role": "user",
      "content": "Solve the new question. Select the most appropriate answer."
    }
  ]
"""
\end{lstlisting}

%% file: latex/sections/appendix/parts/exp_setup.tex
\section{Experimental Setup}
\label{app:experiment_setup}
\setcounter{table}{0}
\begin{table}[H]
\centering
\small
\caption{Experimental setup across model families. "N/A" indicates the parameter count has not been disclosed. Note that embedding and reranker computations were performed on the NVIDIA A40 GPU, with results cached locally.}
\label{tab:models-setup}

\renewcommand\tabularxcolumn[1]{m{#1}}

\begin{tabularx}{\columnwidth}{
    >{\raggedright\arraybackslash}X  
    l                                
    l                                
}
\toprule
\textbf{Model Family} & \textbf{Params} & \textbf{Execution Environment} \\
\midrule
QWen-3 & 8B & NVIDIA A40 (24GB) \\
LLaMA-3.1 & 8B & NVIDIA A40 (24GB) \\
{\deepseek} & 8B & NVIDIA A40 (24GB) \\
MistralSmall & 24B & API endpoint \\
Gemini 2.0 Flash & N/A & API endpoint \\
GPT-4o Mini & N/A & API endpoint \\
GPT-4o & N/A & API endpoint \\
GPT-5 & N/A & API endpoint* \\
\bottomrule
\multicolumn{3}{l}{\footnotesize *Used for a subset of experiments.}
\end{tabularx}
\end{table}

%% file: latex/sections/appendix/parts/synthetic_data.tex
\section{Synthetic Retrieval Pool}
\label{app:synthetic_data_generation}
Below are the prompts that we used to generate the synthetic data that are used as our knowledge base for knowledge and schema retrieval mechanisms. The synthetic data are constructed using GPT-4o via the OpenAI API. Specifically, we provide exact prompts for each of the three \textit{synthetic similarity} levels: {\same}, {\similar}, and {\different}. Note that we intentionally keep all prompts identical to our experiment setup including format, punctuation, and \textbf{typos} to ensure reproducibility.
\subsection{\same}

\begin{lstlisting}[language=Python]
"""
Consider answer({answer}) and the explanation of solving it({explanation}). this question: {question}, along with its 

Please generate a new question that is distinct from the previous question.

You should follow the following criteria:
- New question requires more knowledge than the provided explanation to be used to answer it.
- New question should differ from the given question with a lot of distinctiveness.
- Generate a set of new options with only one of them being the correct option to the new question.
- Provide three incorrect options, which should be similar to the correct answer
- Provide a short explanation on how to solve the new question, and the additional knowledge required to answer the new question.
- Difficulty:
    The new question should be the similar difficulty to the previous question.
    If a student has the knowledge to answer the previous question, they should have partial knowledge to answer the new question.
    However, the new question should require additional knowledge than the given question's scope to be answered.
- Distinctiveness:
    The new question should be distinctive enough to the previous question, that the student require additional knowledge to solve the problem. 
    New question should be unique in its context, and is related to the previous question in a minimal level.
- Output Format:
    {question_format}
"""
\end{lstlisting}

\subsection{\similar}
\begin{lstlisting}[language=Python]
"""
Consider this question: {question}, along with its answer({answer}) and the explanation of solving it({explanation}).

Please give me a slightly different question from this example that test the student's ability to transform their knowledge.

You should follow the following criteria:
- The new question only requires the knowledge provided in the explanation to be used to answer it.
- New question should still differ with a lot of distinctiveness to test student's use of the same knowledge.
- Generate a set of new options with only one of them being the correct option to the new question
- Provide three incorrect options, which should be similar to the correct answer
- Provide a short explanation on how to solve the new question
- Difficulty: 
    The new question should be the similar difficulty to the previous question.
    If a student has the knowledge to answer the previous question, they should have enough knowledge to answer the new question.
- Distinctiveness: 
    The new question should be distinctive enough to the previous question, that the student cannot use the same answer.
    New question should be unique in its context, but still related to the previous question.
- Output Format:
    {question_format}
"""
\end{lstlisting}

\subsection{\different}
\begin{lstlisting}[language=Python]
"""
Consider this question: {question}, along with its answer({answer}) and the explanation of solving it({explanation}).

Please generate a new question that is distinct from the previous question.

You should follow the following criteria:
- New question requires more knowledge than the provided explanation to be used to answer it.
- New question should differ from the given question with a lot of distinctiveness.
- Generate a set of new options with only one of them being the correct option to the new question.
- Provide three incorrect options, which should be similar to the correct answer
- Provide a short explanation on how to solve the new question, and the additional knowledge required to answer the new question.
- Difficulty:
    The new question should be the similar difficulty to the previous question.
    If a student has the knowledge to answer the previous question, they should have partial knowledge to answer the new question.
    However, the new question should require additional knowledge than the given question's scope to be answered.
- Distinctiveness:
    The new question should be distinctive enough to the previous question, that the student require additional knowledge to solve the problem. 
    New question should be unique in its context, and is related to the previous question in a minimal level.
- Output Format:
    {question_format} 
"""
\end{lstlisting}

%% file: latex/sections/appendix/parts/heatmap.tex
\section{Heatmaps for Knowledge Density in Latent Similarity}
\label{app:knowledge_density}
\begin{figure}[htbp]
    \centering
    \begin{minipage}{0.5\textwidth}
        \centering
        \includegraphics[width=\linewidth]{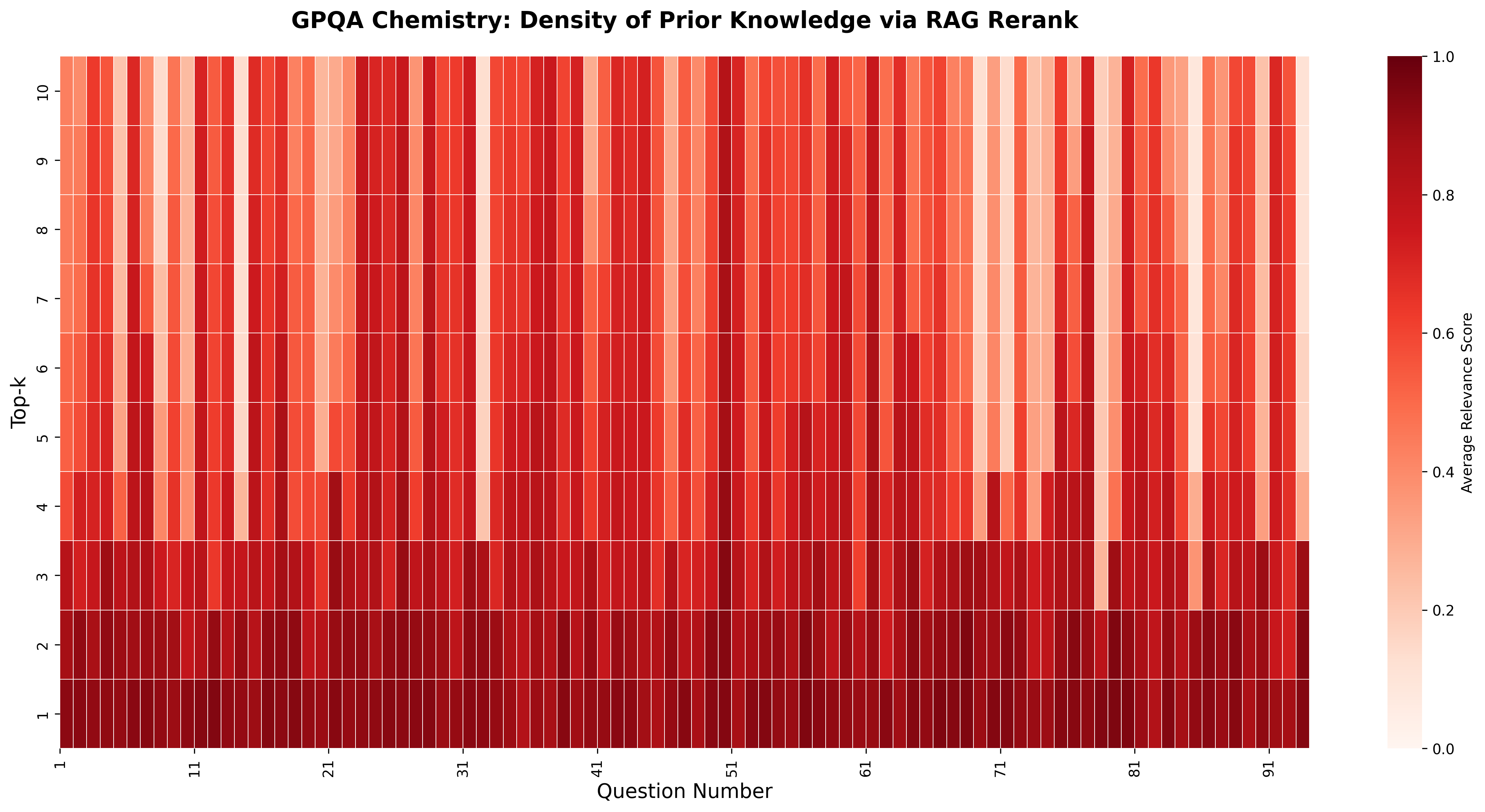} 
    \end{minipage}\hfill
    \begin{minipage}{0.5\textwidth}
        \centering
        \includegraphics[width=\linewidth]{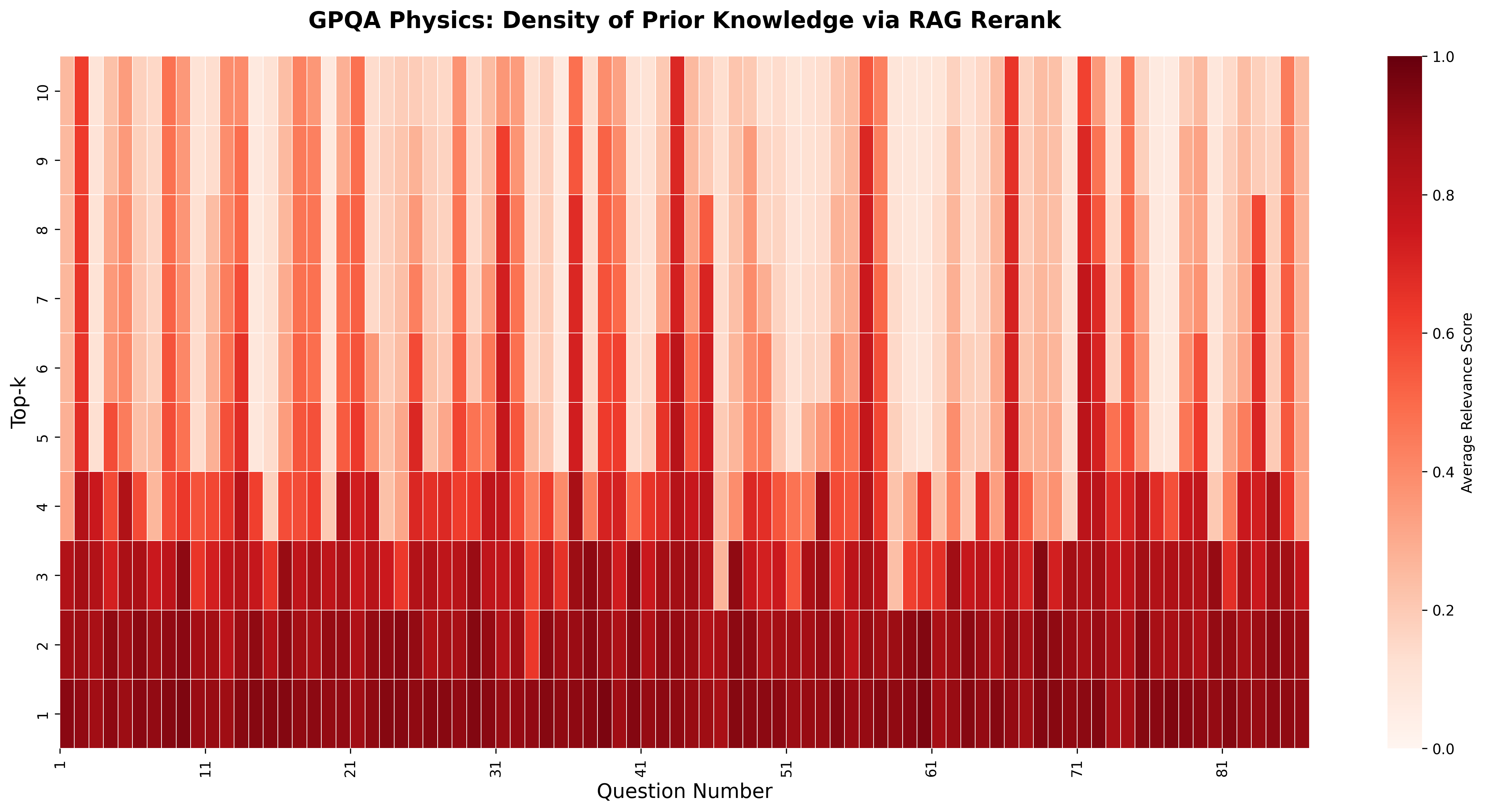}
    \end{minipage}
    
    \caption{The heatmaps for knowledge density for GPQA-Chemistry and GPQA-Physics}
    \label{fig:heatmaps}
\end{figure}

Knowledge density is defined as the similarity between prior knowledge and a target question. Similarity is quantified by converting prior knowledge into mental representations and employing cross-encoder rerankers to compute relevance scores. The knowledge density heatmaps \Cref{fig:heatmaps} visualize marginal relevance scores across top-k retrieved examples for each question, revealing domain-specific patterns in knowledge sparsity. Our analysis demonstrates that chemistry maintains consistently high relevance scores across retrieved examples, while physics exhibits significantly greater sparsity --- with relevance scores decreasing 68\% compared to chemistry's 40\% decrease. These findings underscore that effective schema-based knowledge transfer depends on inherent conceptual coherence within domains rather than universal retrieval mechanisms.

\subsection{Latent Knowledge Density Methodology}
We define the latent knowledge density function $\rho: Q \times K \rightarrow [0, 1]$, where $Q$ represents the set of questions and $K = \{1, 2, ..., k_{max}\}$ denotes retrieval ranks. In our case, $k_{max} := 10$. For each question $q \in Q$ and rank $k \in K$, $\rho(q, k)$ measures the relevance score between $q$ and its $k$-th most similar prior example, computed using {\cohere} cross-encoder architecture. The aggregate density metric $\bar{\rho}(q) = \frac{1}{k_{max}} \sum_{k=1}^{k_{max}} \rho(q, k)$ quantifies overall knowledge availability for question $q$, where higher values indicate robust transfer potential and lower values reveal knowledge sparsity. We define $\sigma \in [0, 1]$ as the hyperparameter of the relevance threshold to quantify what we consider quality knowledge. For this investigation, we arbitrarily set $\sigma := 0.5$.

\subsection{Heatmap Visualization and Interpretation}
Figure \ref{fig:heatmaps} visualizes $\rho(q, k)$ for GPQA-Chemistry and GPQA-Physics datasets as heatmaps, where each cell $(i, j)$ represents $\rho(q_i, k_j)$ with $q_i$ denoting the $i$-th question (horizontal axis) and $k_j \in \{1, 2, ..., k_{max}\}$ the retrieval rank (vertical axis). Colour intensity maps linearly to $\rho$ values, from light shades ($\rho \rightarrow 0$) to dark red ($\rho \rightarrow 1$), symbolizing low to high relevance scores respectively.

Chemistry questions maintain $\rho_{chem}(q, k) > \sigma$ for most $(q, k) \in Q \times K$ pairs even at $k = k_{max}$, indicating persistent semantic similarity with small $\Delta\rho = \rho(q, k) - \rho(q, k+1)$ for all $k \in \{1, 2, ..., k_{max} - 1\}$. Physics exhibits rapid density decay with $\rho_{phys}(q, k) < 0.3$ for many questions by $k = 5$, revealing sparse knowledge availability beyond initial retrievals.

\subsection{Domain Sparsity and Knowledge Space Isolation}
Defining decay rate as $\theta(k) = \frac{\rho(q, k)}{\rho(q, 1)}$, we observe $\bar{\theta}_{chem}(k_{max}) = 0.60$ for chemistry versus $\bar{\theta}_{phys}(k_{max}) = 0.32$ for physics, where the bar notation indicates domain averaging. This represents $\bar{\rho}_{chem}(q, k_{max}) \approx 0.6 \cdot \bar{\rho}_{chem}(q, 1)$ netting a 40\% decrease, compared to $\bar{\rho}_{phys}(q, k_{max}) \approx 0.32 \cdot \bar{\rho}_{phys}(q, 1)$ which yields a 68\% decrease. Chemistry maintains $\bar{\rho}_{chem}(q, k) > \sigma$ throughout all $k \in \{1, 2, ..., k_{max}\}$, while physics shows a domain-specific density gradient $\nabla_k\rho$ approximately twice as steep.

We hypothesize that this disparity stems from the inherent conceptual fragmentation of physics across fields like quantum mechanics and general relativity --- each possessing specialized theoretical vocabularies that create isolated knowledge regions within the latent space. The intercluster density satisfies $\rho(q_i, e_j) \ll \sigma$ when $q_i$ and the episodic trace $e_j$ belong to different physics subdomains. For example, specialized concepts inherent to quantum mechanics, such as quantum entanglement and wave functions, yield $\rho \approx 0$ when matched against general relativity examples, forming disjoint clusters and thus leading to poorer schema activations.

These findings demonstrate that effective schema-based knowledge transfer depends critically on domain density structure $\rho(q, k)$. High-density regimes where $\bar{\rho} > \sigma$ support extensive retrieval sets, while sparse domains with rapidly decaying $\rho(q, k)$ require adaptive mechanisms for $\rho \rightarrow 0$ when $k > k_{threshold}$.

%% file: latex/sections/appendix/parts/schema_template.tex
\section{Domain-Agnostic Schema Template}
\label{app:schema_templates}
{\sysname} integrates seamlessly with empirically proven prompting strategies. While our main experiments primarily utilize {\CoT}, we also evaluated four additional schema templates: Least-to-Most, Step-Back, Deductive-driven, and Facts-driven. Traditional implementations of Least-to-Most and Step-Back often require multiple API calls to build reasoning sequentially. To ensure strict experimental control, we condensed their schema generation into a single API call prior to the final inference step. We call it "\textit{Stage-1}". 
For instance, rather than using multi-turn prompts, our Step-Back template extracts the abstract question, foundational principle, and reasoning outline simultaneously. This isolates the structural benefits of the template while eliminating the confounding variables of multi-turn conversational prompting.

\subsection{Least-to-Most}
\begin{lstlisting}[language=Python,breaklines=true]
"""
You are an expert problem solver using the 'Least-to-Most' prompting strategy.
Your goal is to break down the user's complex problem into a list of simple, sequential sub-questions.

1. Read the problem carefully.
2. Identify the logical steps required to solve it.
3. Output a numbered list of sub-questions in the 'sub_questions' field.
   - The sub-questions MUST be specific to the problem (mention specific numbers or entities).
   - Do NOT solve the sub-questions yet.
   - Ensure the last sub-question asks for the final answer.
"""
\end{lstlisting}

\subsection{Step-Back}
\begin{lstlisting}[language=Python]
"""
You are an expert teacher helping students understand a problem using **Step-Back Prompting**.
Instead of solving the specific details immediately, you must first abstract the problem to a higher level.

Please provide the following:
1. **Step-Back Question**: A generic, high-level question that addresses the underlying concept, theory, or physics law behind the specific problem. This question should be abstract enough to apply to a class of similar problems.
2. **Principle**: The fundamental principle, formula, or fact required to answer that step-back question.
3. **Reasoning Steps**: A concise, step-by-step outline of how to apply that principle to solve the specific problem.

**Your Schema must not include the final answer or any candidate option.**
**Keep the steps concise. Use bullet points.**
"""
\end{lstlisting}

\subsection{Deductive-driven}
\begin{lstlisting}[language=Python]
"""
You are a teacher helping students understand a problem. Drawing on schema theory from cognitive psychology, provide a high-level abstraction (schema) of the problem to guide your students. Your schema should include the following components:

Broad Category:
Identify the overarching subject and general category to which the problem belongs.

Refinement:
Describe further details or specific aspects that narrow down the broad category.

Specific Scope:
Define the precise focus or context of the problem within the refined category.

Goal:
Clearly state the objective or intended outcome of solving the problem.
"""
\end{lstlisting}

\subsection{Facts-driven}
\begin{lstlisting}[language=Python]
"""
You are a teacher helping students understand a problem.
Drawing on schema theory from cognitive psychology, provide a high-level abstraction (schema) of the problem to guide your students. Your schema should include a list of the general facts in a paragraph. The facts should not mention the problem itself. One sentence per fact.
"""
\end{lstlisting}

%% file: latex/sections/appendix/parts/vision.tex
\section{Experiment with a Multi-Modal dataset}

\ifthenelse{\equal{\papertype}{preprint}}{
    \label{app:vision}
    \begin{figure}[h!]
      \centering
      \includegraphics[width=0.45\linewidth]{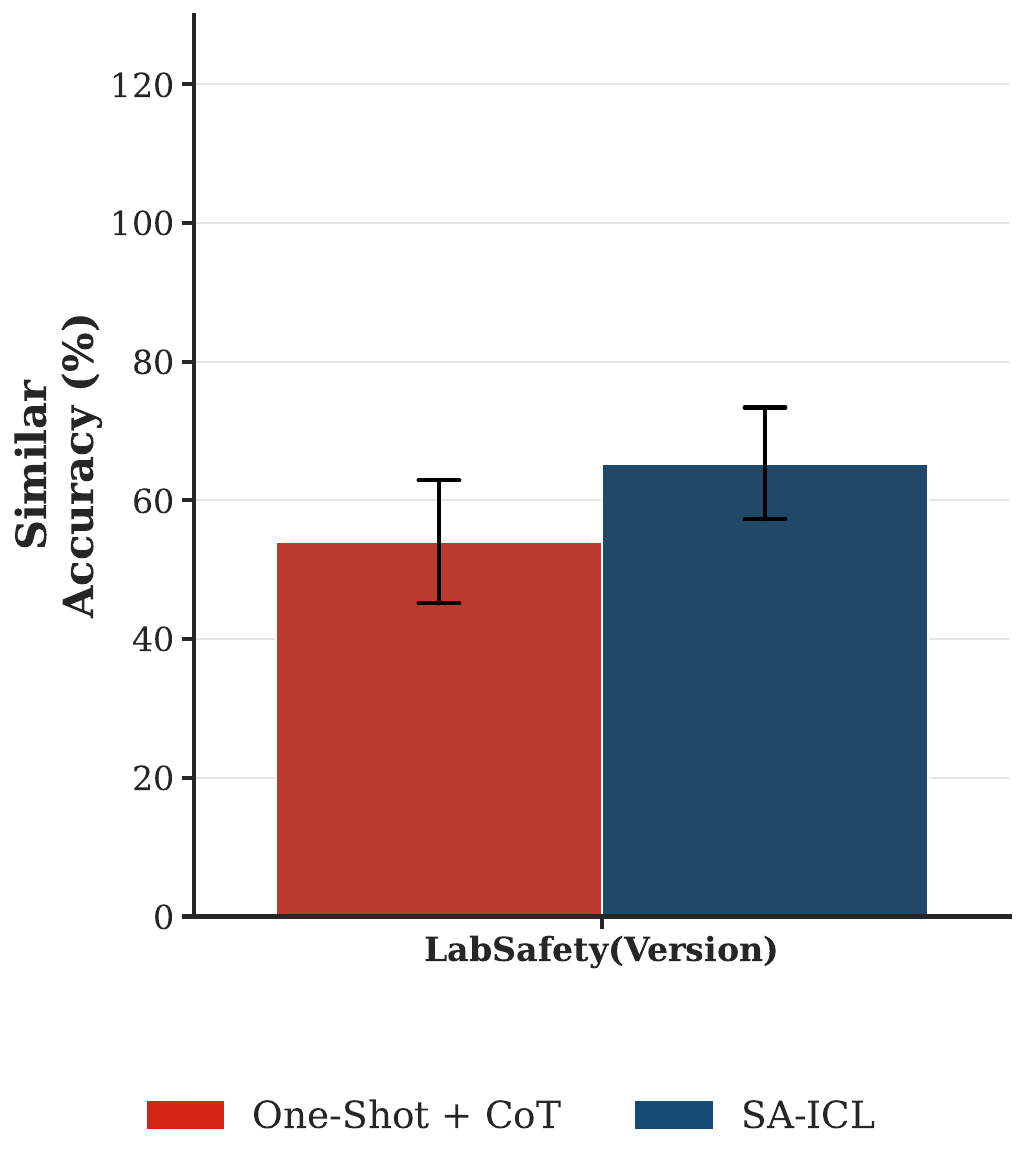}
      \caption{The performances of Gemini 2.0-Flash on the LabSafety dataset.}
      \label{fig:vision}
    \end{figure}
}{
    \label{app:vision}
    \begin{figure}[h!]
      \centering
      \includegraphics[width=\linewidth]{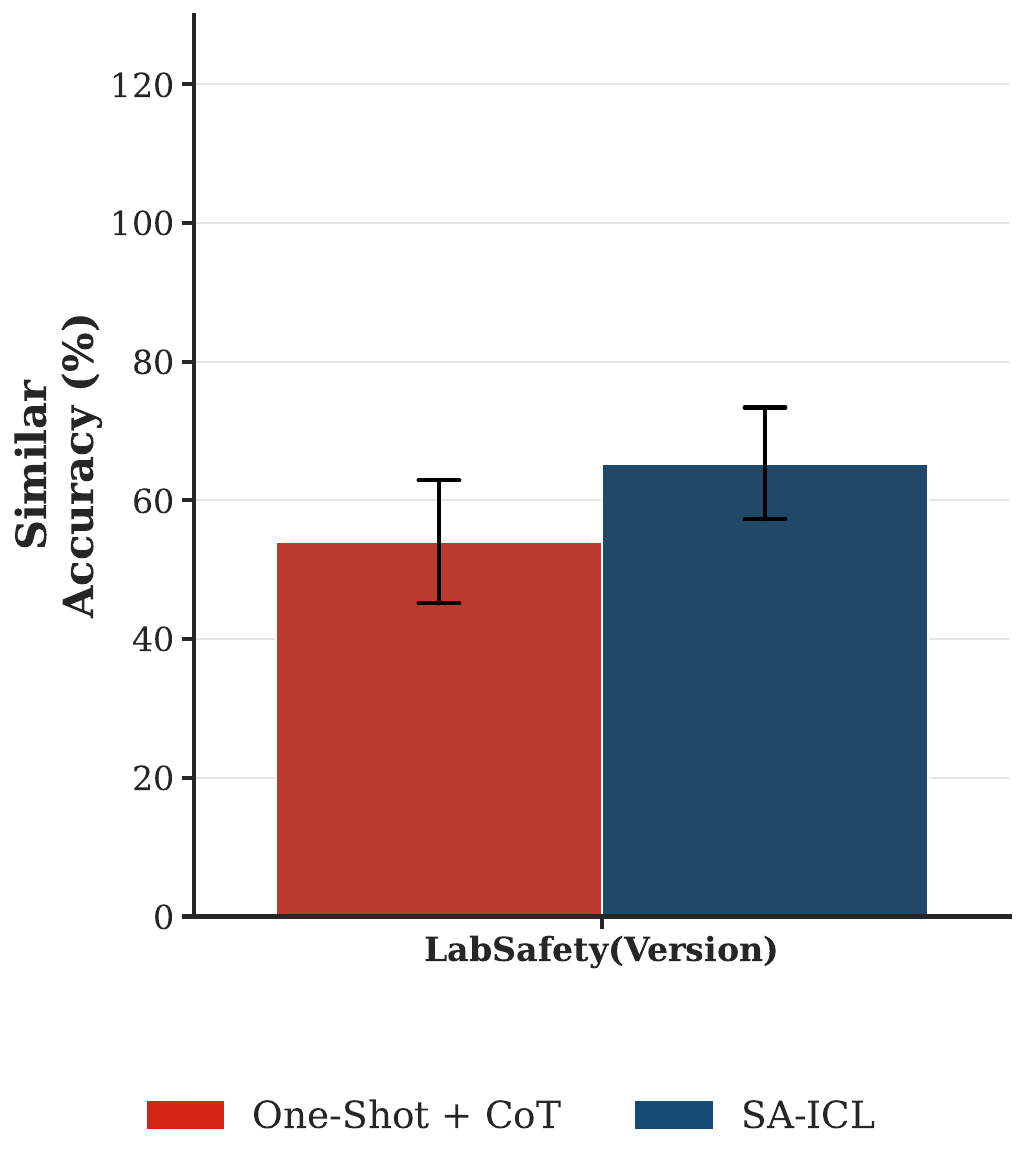}
      \caption{The performances of Gemini 2.0-Flash on the LabSafety dataset.}
      \label{fig:vision}
    \end{figure}
}

From \Cref{fig:vision}, we saw that by using {\sysname}, Gemini 2.0-Flash utilized the examples better. This result indicates the potential of {\LLMsFull} to interpret the visualizations through the lens of {\SchemaTheory}

For this experiment, due to the difficulty of generating synthetic examples with images, we used one question from a specific category as the example for other questions in the same category.

%% file: latex/sections/appendix/parts/full_exp_analysis.tex
\section{Extra Main Experiment Result Analysis}
\label{app:extra}
\subsection{Effectiveness of {\sysname} per model}
\input{latex/sections/Results/pvalue_per_model}
\Cref{tab:p_value_per_model} shows that {\sysname} is a remarkably effective {\ICL} approach for five out of seven models in the three domains: {\science}, {\stat}, {\finance}.

\subsection{Effectiveness of {\sysname} when the knowledge level is low}
When the knowledge level is low, or when {\LLMs} were given an example that was different from the question, \Cref{fig:different,fig:trend} shows that {\sysname} became weaker, despite having a global improvement of +3.5\%, compared to {\OneShotCoT}. It is worth pointing out that {\different} is the only knowledge level for which we did not find a significant effectiveness of {\sysname} in our main experiments.

\subsection{GPT-5's Performance on Humanity's Last Exam}
\label{app:gpt5}
While we focused mostly on smaller models, we believe it's worth showing the results of a larger model on an even more challenging dataset, to better justify the usage of {\sysname} beyond smaller models. Here, we reported the result of using {\sysname} with GPT-5 on HLE subsets in \Cref{fig:gpt5}.
\ifthenelse{\equal{\papertype}{preprint}}{
    \begin{figure}[h!]
      \centering
      \includegraphics[width=0.4\linewidth]{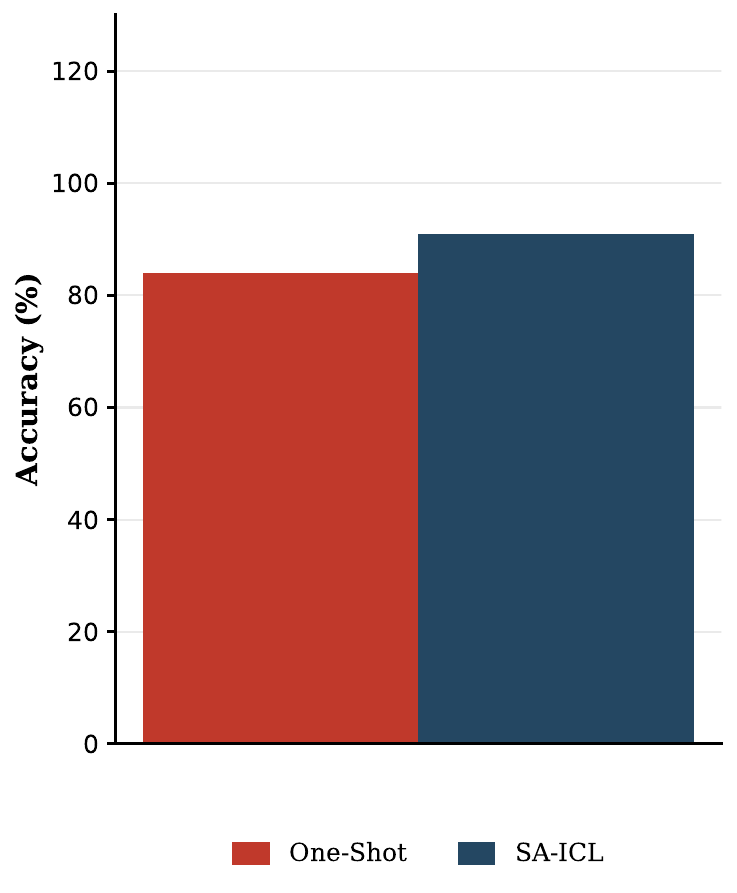}
      \caption{Performance of GPT-5 on the chemistry, physics, biology multi-choice questions from the Humanity's Last Exam (HLE) benchmark dataset \citep{phan2025humanity} (N=191, we filtered out questions with images), when provided with the {\same} examples. We generated the synthetic data as well using the same procedures described in \Cref{app:synthetic_data_generation}.}
      \label{fig:gpt5}
    \end{figure}
}{
    \begin{figure}[h!]
      \centering
      \includegraphics[width=\linewidth]{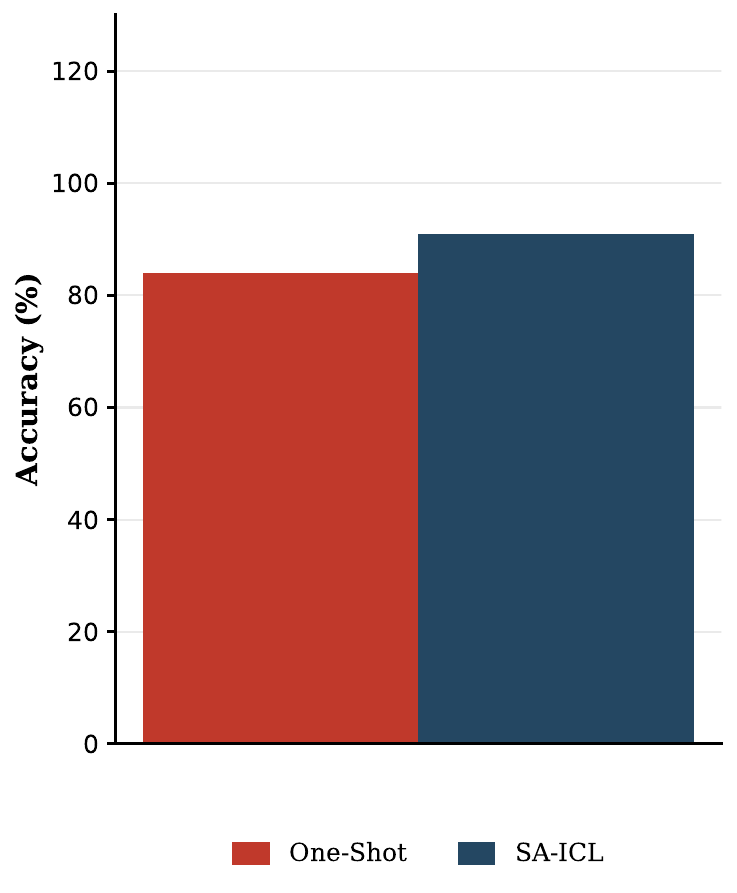}
      \caption{Performance of GPT-5 on the chemistry, physics, biology multi-choice questions from the Humanity's Last Exam (HLE) benchmark dataset \citep{phan2025humanity} (N=191, we filtered out questions with images), when provided with the {\same} examples. We generated the synthetic data as well using the same procedures described in \Cref{app:synthetic_data_generation}.}
      \label{fig:gpt5}
    \end{figure}
}
When GPT-5 was introduced, we used the model to run the same experiments. However, we realized that when provided with {\same} examples, GPT-5 reached near 100\% accuracy in the GPQA dataset, eliminating the need for explicit reasoning modules introduced by {\sysname}. Therefore, we used a more challenging dataset, Humanity's Last Exam (HLE), to repeat the same experiments using {\same} examples, and saw that {\sysname} results in a 7\% improvement in accuracy compared to standard {\oneshot}. This gain is particularly notable given the already strong baseline accuracy of 84\%.

Note that this subset of HLE includes 136 biology and medicine questions, 34 physics questions, and 21 chemistry questions. We acknowledge that most of the questions are in the biology and medicine sub-categories, which is a domain that was not tested in the main experiments shown in \Cref{tab:full_results}.

%% file: latex/sections/Results/pvalue_per_model.tex
\ifthenelse{\equal{\papertype}{preprint}}{
    \begin{table}[htpb]
        \centering
        \caption{Performance Impact Analysis by Model, comparing Reasoning vs. Schema Impact across different LLMs. There are no significant improvements by only having {\CoT} in the context, but {\sysname} would give significant improvements in five out of the seven models we tested with.}
        \label{tab:p_value_per_model}
        \resizebox{0.75\linewidth}{!}{%
            \begin{tabular}{l c c c l}
                \toprule
                \textbf{Metric} & \textbf{Baseline} & \textbf{Final} & \textbf{Change ($\Delta$)} & \textbf{$p$-value} \\
                \midrule
                \multicolumn{5}{l}{\textbf{Gemini}} \\
                \hspace{3mm} Reasoning Impact & 39.0\% & 41.3\% & +2.3\% & 0.2882 (ns) \\
                \hspace{3mm} Schema Impact    & 41.3\% & 47.4\% & +6.2\% & 0.0050 (**) \\
                \midrule
                \multicolumn{5}{l}{\textbf{GPT4o}} \\
                \hspace{3mm} Reasoning Impact & 51.0\% & 51.3\% & +0.3\% & 0.8947 (ns) \\
                \hspace{3mm} Schema Impact    & 51.3\% & 52.2\% & +1.0\% & 0.6588 (ns) \\
                \midrule
                \multicolumn{5}{l}{\textbf{GPT4oMini}} \\
                \hspace{3mm} Reasoning Impact & 44.7\% & 43.6\% & -1.2\% & 0.5939 (ns) \\
                \hspace{3mm} Schema Impact    & 43.6\% & 48.1\% & +4.5\% & 0.0390 (*) \\
                \midrule
                \multicolumn{5}{l}{\textbf{Llama3}} \\
                \hspace{3mm} Reasoning Impact & 31.1\% & 33.2\% & +2.1\% & 0.2994 (ns) \\
                \hspace{3mm} Schema Impact    & 33.2\% & 39.3\% & +6.1\% & 0.0044 (**) \\
                \midrule
                \multicolumn{5}{l}{\textbf{Qwen3}} \\
                \hspace{3mm} Reasoning Impact & 46.9\% & 48.4\% & +1.5\% & 0.5082 (ns) \\
                \hspace{3mm} Schema Impact    & 48.4\% & 50.7\% & +2.3\% & 0.2902 (ns) \\
                \midrule
                \multicolumn{5}{l}{\textbf{Deepseek}} \\
                \hspace{3mm} Reasoning Impact & 45.6\% & 43.2\% & -2.4\% & 0.2672 (ns) \\
                \hspace{3mm} Schema Impact    & 43.2\% & 51.6\% & +8.4\% & 0.0001 (***) \\
                \midrule
                \multicolumn{5}{l}{\textbf{MistralSmall}} \\
                \hspace{3mm} Reasoning Impact & 45.3\% & 44.4\% & -0.9\% & 0.6923 (ns) \\
                \hspace{3mm} Schema Impact    & 44.4\% & 51.5\% & +7.1\% & 0.0017 (**) \\
                \bottomrule
            \end{tabular}%
        }
    \end{table}
}{
    \begin{table}[htpb]
        \centering
        \caption{Performance Impact Analysis by Model, comparing Reasoning vs. Schema Impact across different LLMs. There are no significant improvements by only having {\CoT} in the context, but {\sysname} would give significant improvements in five out of the seven models we tested with.}
        \label{tab:p_value_per_model}
        \begin{adjustbox}{max width=\linewidth}
            \begin{tabular}{l c c c l}
                \toprule
                \textbf{Metric} & \textbf{Baseline} & \textbf{Final} & \textbf{Change ($\Delta$)} & \textbf{$p$-value} \\
                \midrule
                \multicolumn{5}{l}{\textbf{Gemini}} \\
                \hspace{3mm} Reasoning Impact & 39.0\% & 41.3\% & +2.3\% & 0.2882 (ns) \\
                \hspace{3mm} Schema Impact    & 41.3\% & 47.4\% & +6.2\% & 0.0050 (**) \\
                \midrule
                \multicolumn{5}{l}{\textbf{GPT4o}} \\
                \hspace{3mm} Reasoning Impact & 51.0\% & 51.3\% & +0.3\% & 0.8947 (ns) \\
                \hspace{3mm} Schema Impact    & 51.3\% & 52.2\% & +1.0\% & 0.6588 (ns) \\
                \midrule
                \multicolumn{5}{l}{\textbf{GPT4oMini}} \\
                \hspace{3mm} Reasoning Impact & 44.7\% & 43.6\% & -1.2\% & 0.5939 (ns) \\
                \hspace{3mm} Schema Impact    & 43.6\% & 48.1\% & +4.5\% & 0.0390 (*) \\
                \midrule
                \multicolumn{5}{l}{\textbf{Llama3}} \\
                \hspace{3mm} Reasoning Impact & 31.1\% & 33.2\% & +2.1\% & 0.2994 (ns) \\
                \hspace{3mm} Schema Impact    & 33.2\% & 39.3\% & +6.1\% & 0.0044 (**) \\
                \midrule
                \multicolumn{5}{l}{\textbf{Qwen3}} \\
                \hspace{3mm} Reasoning Impact & 46.9\% & 48.4\% & +1.5\% & 0.5082 (ns) \\
                \hspace{3mm} Schema Impact    & 48.4\% & 50.7\% & +2.3\% & 0.2902 (ns) \\
                \midrule
                \multicolumn{5}{l}{\textbf{Deepseek}} \\
                \hspace{3mm} Reasoning Impact & 45.6\% & 43.2\% & -2.4\% & 0.2672 (ns) \\
                \hspace{3mm} Schema Impact    & 43.2\% & 51.6\% & +8.4\% & 0.0001 (***) \\
                \midrule
                \multicolumn{5}{l}{\textbf{MistralSmall}} \\
                \hspace{3mm} Reasoning Impact & 45.3\% & 44.4\% & -0.9\% & 0.6923 (ns) \\
                \hspace{3mm} Schema Impact    & 44.4\% & 51.5\% & +7.1\% & 0.0017 (**) \\
                \bottomrule
            \end{tabular}%
        \end{adjustbox}
    \end{table}
}

\begin{figure*}[t]
  \centering
  \includegraphics[width=\linewidth]{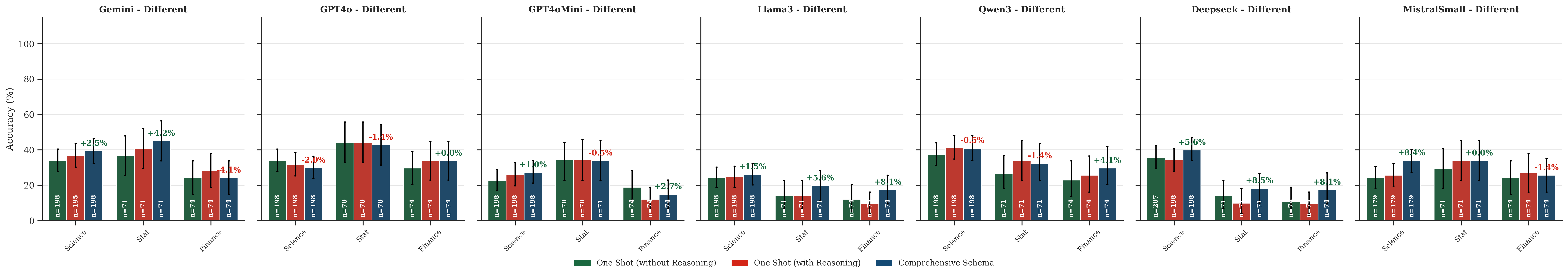}
  \caption{The {\LLMs}' performances when the example is different from the target questions.}
  \label{fig:different}
\end{figure*}

%% file: latex/sections/appendix/parts/history.tex
\section{Robustness against an irrelevant example}
\label{app:history}

\begin{figure}[h!]
  \centering
  \includegraphics[width=\linewidth]{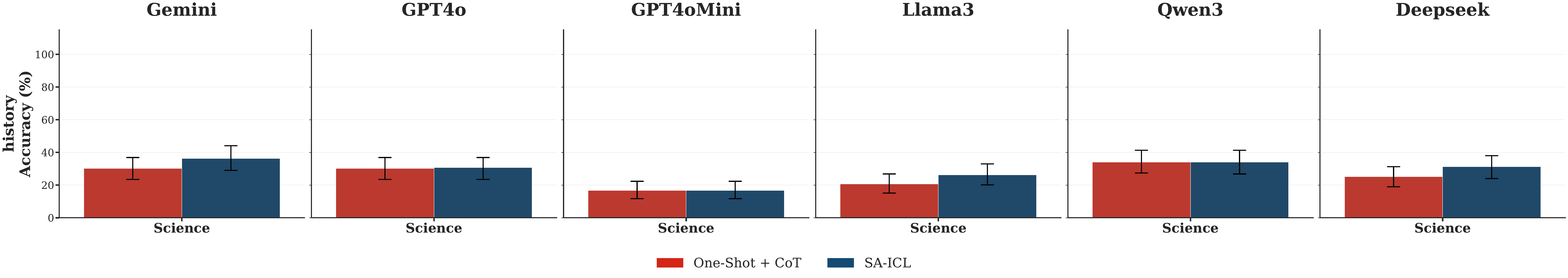}
  \caption{The performances of six models when they were given a biology question as the in-context example to solve Chemistry and Physics questions.}
  \label{fig:history}
\end{figure}
We tested the susceptibility to Schema Disruption via out-of-domain examples. To evaluate whether irrelevant context misleads the model, a Biology question was introduced as a few-shot exemplar during Chemistry and Physics inference. Despite both conditions perform worse, {\sysname} suffers from a lighter decrease in the performances as shown in \Cref{fig:history}.

%% file: latex/sections/appendix/parts/semantic_shift.tex
\section{Schema Shifting by models}
\label{app:schema_shift}
\ifthenelse{\equal{\papertype}{preprint}}{
    \begin{figure}[H]
      \centering
      \includegraphics[width=0.7\linewidth]{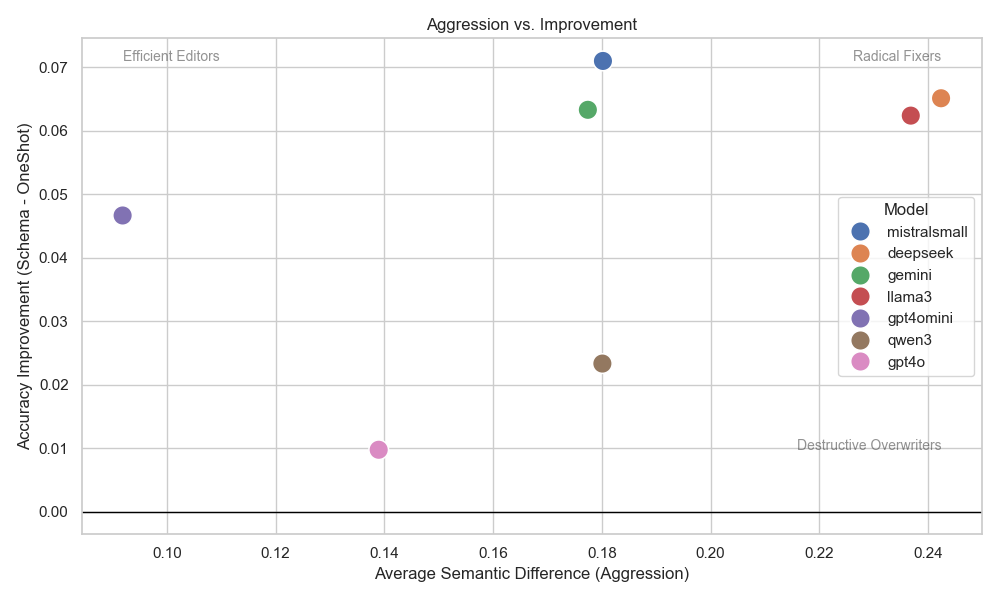}
      \caption{For questions that {\OneShotCoT} answered incorrectly — indicating a faulty latent schema — {\sysname} achieved a repair rate of up to 7\%. While different models exhibited varying degrees of schema shift, a consistent trend emerges: greater semantic differences between the latent and activated schemas are associated with a higher probability of repair.}
      \label{fig:schema_shift_per_model}
    \end{figure}
}{
    \begin{figure}[H]
      \centering
      \includegraphics[width=\linewidth]{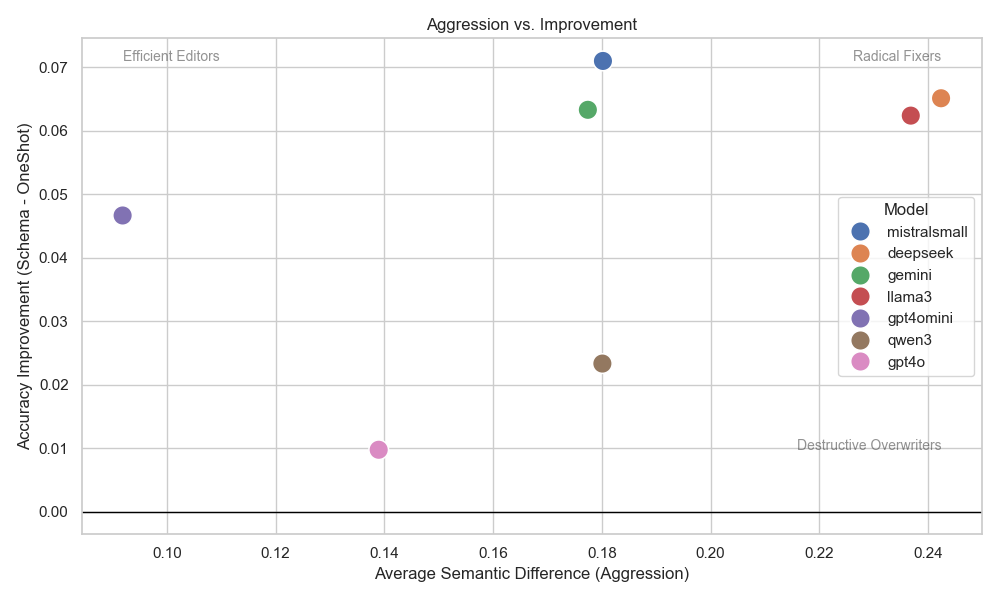}
      \caption{For questions that {\OneShotCoT} answered incorrectly — indicating a faulty latent schema — {\sysname} achieved a repair rate of up to 7\%. While different models exhibited varying degrees of schema shift, a consistent trend emerges: greater semantic differences between the latent and activated schemas are associated with a higher probability of repair.}
      \label{fig:schema_shift_per_model}
    \end{figure}
}